\documentclass[11pt]{article}
\usepackage[a4paper,margin=2.5cm]{geometry}
\usepackage[T1]{fontenc}
\usepackage{lmodern,microtype,amsmath,amssymb,booktabs,array,graphicx,xcolor,multirow}
\usepackage[numbers,sort&compress]{natbib}
\usepackage{tikz,pifont,fontawesome5}
\usetikzlibrary{arrows.meta,positioning,calc,shadows}
\usepackage[hidelinks]{hyperref}

\IfFileExists{generated/numbers.tex}{
\newcommand{\FullMaxDiffBacc}{0.006}
\newcommand{\FullMaxDiffEce}{0.030}

\newcommand{\HOneADiffCONCH}{-0.199}

\newcommand{\HOneADiffSFT}{-0.075}
\newcommand{\HOneAHiCONCH}{-0.158}

\newcommand{\HOneAHiSFT}{-0.064}
\newcommand{\HOneALoCONCH}{-0.237}

\newcommand{\HOneALoSFT}{-0.088}

\newcommand{\HOneBDiffCONCH}{+0.196}

\newcommand{\HOneBDiffSFT}{+0.049}
\newcommand{\HOneBHiCONCH}{+0.218}

\newcommand{\HOneBHiSFT}{+0.063}
\newcommand{\HOneBLoCONCH}{+0.161}

\newcommand{\HOneBLoSFT}{+0.038}

\newcommand{\NClassesHepato}{7}
\newcommand{\NClassesHiesd}{10}
\newcommand{\NOrganCandidates}{10}
\newcommand{\NResamples}{10{,}000}
\newcommand{\NSeeds}{3}

\newcommand{\NTestCap}{5{,}000}

\newcommand{\NTrainStates}{40{,}000}
\newcommand{\NValStates}{2{,}100}
\newcommand{\ResCONCHCleanBacc}{0.345}

\newcommand{\ResCONCHCleanEce}{0.403}
\newcommand{\ResCONCHCleanEceT}{0.086}

\newcommand{\ResCONCHCleanNllT}{1.597}

\newcommand{\ResCONCHUnseenBacc}{0.328}

\newcommand{\ResIJCleanBacc}{0.288}

\newcommand{\ResOMJCleanBacc}{0.213}

\newcommand{\ResPJCleanBacc}{0.541}
\newcommand{\ResPJCleanBaccSD}{0.015}

\newcommand{\ResPJCleanEce}{0.204}
\newcommand{\ResPJCleanEceSD}{0.002}
\newcommand{\ResPJCleanEceT}{0.154}
\newcommand{\ResPJCleanNll}{1.586}
\newcommand{\ResPJCleanNllT}{1.399}
\newcommand{\ResPJInDomBacc}{0.692}

\newcommand{\ResPJInDomEce}{0.108}

\newcommand{\ResPJSeenBacc}{0.781}

\newcommand{\ResPJUnseenBacc}{0.390}

\newcommand{\ResPJUnseenEce}{0.301}

\newcommand{\ResProbeInDomBacc}{0.677}

\newcommand{\ResProbeInDomEce}{0.074}

\newcommand{\ResProbeSeenBacc}{0.905}

\newcommand{\ResProbeUnseenBacc}{0.772}

\newcommand{\ResSFTCleanBacc}{0.492}

\newcommand{\ResSFTCleanEce}{0.279}

\newcommand{\ResSFTCleanEceT}{0.211}

\newcommand{\ResSFTCleanNllT}{1.795}
\newcommand{\ResSFTInDomBacc}{0.707}

\newcommand{\ResSFTSeenBacc}{0.810}

\newcommand{\ResSFTUnseenBacc}{0.277}

\newcommand{\TmpCONCHValNll}{1.769}

\newcommand{\TmpPJCrossNll}{1.781}

\newcommand{\TmpPJOracleNll}{1.713}

\newcommand{\TmpPJValNll}{2.079}

}{}
\IfFileExists{generated/numbers_a3hepato.tex}{

\newcommand{\AThreeHepNoHepHOneBDiffSFT}{+0.033}

\newcommand{\AThreeHepNoHepHOneBHiSFT}{+0.052}

\newcommand{\AThreeHepNoHepHOneBLoSFT}{+0.018}

\newcommand{\AThreeHepProbeBlockFiftyBacc}{0.967}

\newcommand{\AThreeHepProbeBlockTwoHundredBacc}{0.959}

\newcommand{\AThreeHepProbeImageBacc}{0.971}

}{}
\IfFileExists{generated/numbers_a3.tex}{
\newcommand{\AThreeOMJNineSixCleanBacc}{0.207}

\newcommand{\AThreePermOMJFlip}{18.9}

\newcommand{\AThreePermPJBaccMax}{0.549}
\newcommand{\AThreePermPJBaccMin}{0.542}

\newcommand{\AThreePermPJFlip}{7.3}
\newcommand{\AThreePermPJN}{10}

\newcommand{\AThreeSFTLenCleanBacc}{0.492}
\newcommand{\AThreeSFTLenCleanEce}{0.201}
\newcommand{\AThreeSFTLenCleanNll}{1.508}

\newcommand{\AThreeSFTLenInDomEce}{0.293}

\newcommand{\AThreeSFTLenUnseenBacc}{0.289}
\newcommand{\AThreeSFTLenUnseenEce}{0.110}

}{}
\IfFileExists{generated/numbers_audit.tex}{
\newcommand{\AuditEosMaxDp}{2.4\times10^{-6}}
\newcommand{\AuditEosMinProb}{0.9993}
\newcommand{\AuditEosShareExact}{99}
\newcommand{\AuditRerunMaxDp}{0.12}
\newcommand{\AuditRerunMaxMacroDiff}{0.002}
\newcommand{\AuditRerunMinAgree}{97.7}

\newcommand{\AuditSelRuns}{12}

\newcommand{\SFTExamplesPerPass}{170{,}390}
\newcommand{\SFTPassesAtCkpt}{0.94}
}{}

\title{PathJev: Decision-Native Vision--Language Modelling for Histopathology}
\author{Cristian Tommasino\\
  DIETI, University of Naples Federico II\\
  Naples, Italy\\
  \texttt{cristian.tommasino@unina.it}
  \and
  Antonio M.\ Rinaldi\\
  DIETI, University of Naples Federico II\\
  Naples, Italy}
\date{}

\begin{document}
\maketitle

\begin{abstract}
Vision--language models used for decisions in histopathology rarely return calibrated probabilities over the candidates a user cares about. Decision-native models write a question and a candidate set chosen at run time into the input and return a normalised distribution from a small decision head in one forward pass. We build PathJev-P, a decision-native model for histopathology: MedGemma-1.5-4B with LoRA adapters and a decision head, trained on \NTrainStates{} patch states from three public collections whose expert labels are turned into typed questions with organ-specific candidate sets and a ``none of these'' option. In a comparison over \NSeeds{} seeds with an analysis plan fixed before testing, on test sets without documented overlap with the training data of the backbone, PathJev-P is more accurate than a generative model trained with the same backbone, data and selection rule, the zero-shot pathology model CONCH and a general medical decision model (balanced accuracy \ResPJCleanBacc{} against \ResSFTCleanBacc{}, \ResCONCHCleanBacc{} and \ResOMJCleanBacc{}); its advantage over the generative model comes from two organs absent from training. The calibration advantage depends on how the generative model's probabilities are read out. With the summed candidate log-likelihood used in previous work, PathJev-P has the lower calibration error (ECE \ResPJCleanEce{} against \ResSFTCleanEce{}); with a length-normalised read-out of the same generative model, examined post hoc, the two have similar mean calibration error (\AThreeSFTLenCleanEce{}) and the generative model is better calibrated on the unseen organs, where PathJev-P remains over-confident. A supervised linear probe on a pathology foundation model is better calibrated in every setting. In a qualitative whole-slide demonstration, PathJev-P labels unsupervised tissue clusters with abstention and separates carcinoma from other tissue while running the model on a fraction of the tiles.
\end{abstract}

\section{Introduction}
\label{sec:intro}

Most uses of a vision--language model in pathology end in a decision: which tissue is this, is carcinoma present, which grade, which of these regions needs a closer look. What a downstream rule needs from the model is a probability distribution over the candidates the user cares about, including the option that none of them applies, and probabilities that can be trusted when the tissue comes from an organ or a laboratory the model has not seen. Contrastive pathology models return similarity scores whose scale is arbitrary~\citep{lu2024conch}, and generative assistants return free text whose confidence is a token probability~\citep{lu2024pathchat}; calibration studies of medical vision--language models find the problem particularly severe in pathology~\citep{sharifdeen2026mvcbench,khanmohammadi2026calibratedtriage}. The need is sharpest in whole-slide analysis, where unsupervised tissue clusters are still named by hand~\citep{claudioquiros2024hpl} or with uncalibrated keyword scores~\citep{nguyen2025vleer}.

Decision-native, or ``System One'', models offer a different output channel~\citep{almeida2026jev}: the question and a candidate set chosen at run time are written into the input, and a small decision head returns a normalised distribution over the candidates in one forward pass. The idea has spread quickly to general benchmarks, multimodal models and medical imaging~\citep{deusser2026jevbenchmark,zhu2026interndecision,tang2026omnimedjev}, but three questions matter for pathology and remain open. Does decision-native training carry over to histology, where the vocabulary is organ-specific and the morphology is unlike that of other medical images, and does it generalise to organs absent from training? Are the calibration gains reported over generative models a property of the decision head, or of the way generative probabilities are reconstructed for the comparison? And can such a model name tissue clusters in whole slides, abstaining when it is unsure?

We address these questions with PathJev-P, a decision-native model for histopathology built on MedGemma-1.5-4B~\citep{sellergren2026medgemmaonefive}. We turn the expert labels of three public collections into \NTrainStates{} patch states with typed questions (one of $K$ classes, yes or no, ordered levels), organ-specific candidate sets and a ``none of these'' option, and we evaluate on breast and prostate sets from the training sources and on liver and stomach, two organs absent from training, chosen to have no documented overlap with the backbone's training data. The comparison includes a generative model trained on the same states with the same backbone, zero-shot pathology vision--language models, a general medical decision model and a supervised probe on a pathology foundation model, with an analysis plan fixed before testing.

PathJev-P is more accurate than the generative model, the zero-shot models and the general medical decision model, and its advantage over the generative model lies almost entirely on the new organs. Its calibration advantage, in contrast, depends on the comparator's read-out: it holds when the generative model's candidate probabilities are computed from summed log-likelihoods, as in previous work, but not when the same log-likelihoods are normalised by length. On new organs PathJev-P remains over-confident, and a supervised linear probe is better calibrated in every setting and far more accurate when given labels from the test domain. In a whole-slide demonstration, PathJev-P labels unsupervised tissue clusters with abstention while running on a fraction of the tiles.

\paragraph{Contributions.}
\begin{itemize}
  \item \textbf{PathJev}, a decision-native vision--language model for histopathology. Given an image, a question and a candidate set chosen at run time, it returns a normalised distribution over the candidates, including ``none of these'', in one forward pass, for three question types: one of $K$ classes (Choice), yes or no (Noul) and ordered levels (Score).
  \item \textbf{A construction of decision data from pathology annotations.} Expert labels of heterogeneous sources are turned deterministically into typed questions with organ-specific candidate sets, without generated labels, and split by patient or slide with whole organs held out from training, so that the model can be tested on organs it has never seen.
  \item \textbf{Cluster-level labelling of whole slides.} Tiles are embedded with a pathology foundation model and grouped by a mixture model with soft memberships; PathJev labels each cluster from its most typical tiles, abstains on ambiguous clusters, and its cluster probabilities are propagated to every tile to form probability maps.
  \item \textbf{An evaluation protocol for decision-native models} that compares them with a generative model trained on the same data under several read-outs of its probabilities, candidate orders and image-free inputs, on test sets without documented overlap with the backbone's training data.
\end{itemize}
Code, question templates, split manifests and the analysis plan are available at \url{https://github.com/CosmoIknosLab/PathJev}, and the model weights and a demonstration at \url{https://huggingface.co/Cristiandieti/PathJev-P}.

\section{Related work}
\label{sec:related}

In this section we review decision-native models, pathology vision--language models, the automatic labelling of tissue clusters in whole slides and the calibration of medical vision--language models, and we position PathJev with respect to each line of work.

\paragraph{Decision-native (``System One'') models.} System One models return a typed, normalised distribution over a candidate set supplied at run time instead of generated text~\citep{almeida2026jev}. Since September 2026 the idea has been benchmarked on general tasks~\citep{deusser2026jevbenchmark}, formalised~\citep{li2026jeva}, extended to bi-encoders and smaller multimodal models~\citep{cheng2026metaencoder,zhu2026interndecision} and to visual choice tasks~\citep{zhou2026pixeljev,yu2026visualjev}, and approximated without training by reading the letter logits of instruction-tuned models~\citep{geuter2026innerjev}, which we use as a baseline. Visual Jev~\citep{yu2026visualjev} reads answer probabilities from the language-model head of a vision--language model and reports no consistent accuracy advantage for a typed decision head on general vision benchmarks, a reminder that the effect of the head is hard to separate from that of training. OmniMed-Jev~\citep{tang2026omnimedjev} adapted the recipe to medical imaging and reported calibration gains over a matched generative baseline; its histopathology coverage is limited to PathMNIST. Medical evaluations of System One models are otherwise text-only~\citep{madridgarcia2026jevmedicine,du2026systemonemedicine} or zero-shot on ultrasound~\citep{minh2026breastus}, and \citet{wienholt2026singlepass} read single-pass yes/no decision scores from MedGemma and Gemma-3 without training. Risk-controlled thresholds for such scores are studied in~\citep{messoudi2026riskbudget}. To our knowledge no decision-native model has been trained or evaluated for histopathology with contamination-aware test sets.

\paragraph{Pathology vision--language models.} Contrastive models such as PLIP~\citep{huang2023plip} and CONCH~\citep{lu2024conch} enable zero-shot classification, zero-shot segmentation maps~\citep{moreno2025zeus}, text-based interpretable embeddings~\citep{nguyen2024tqx} and concept-based explanations of slide classifiers~\citep{lazard2026scope}; general VLMs have also served as zero-annotation tissue segmenters~\citep{jain2026zeroannotation}. Contrastive scores are similarities whose softmax temperature is arbitrary, and they offer no explicit ``none of these'' answer. Generative assistants such as PathChat~\citep{lu2024pathchat} answer in free text; their confidence is a token probability, not a decision probability. Calibration studies of medical VLMs report that miscalibration is particularly severe in pathology~\citep{sharifdeen2026mvcbench,khanmohammadi2026calibratedtriage}, and that explicit abstention options help~\citep{machcha2026medabstain}.

\paragraph{Labelling unsupervised tissue clusters.} Clustering tile embeddings recovers histomorphological phenotypes~\citep{claudioquiros2024hpl}, but naming the clusters is manual, done through keyword similarity~\citep{nguyen2025vleer}, or delegated to generative descriptions of cluster representatives~\citep{ding2025titan,gao2026alpaca}; none outputs calibrated, abstaining cluster labels. Our whole-slide demonstration is a first step in this direction; its systematic evaluation is future work.

\paragraph{Calibration.} We use ECE and adaptive ECE~\citep{guo2017calibration,nixon2019measuring}, proper scoring rules, and temperature scaling~\citep{guo2017calibration}; uncertainty thresholds for confident histopathology predictions were advocated by~\citet{dolezal2022uncertainty}.

\section{Method}
\label{sec:method}

\begin{figure*}[t]
\centering
\includegraphics[width=\textwidth]{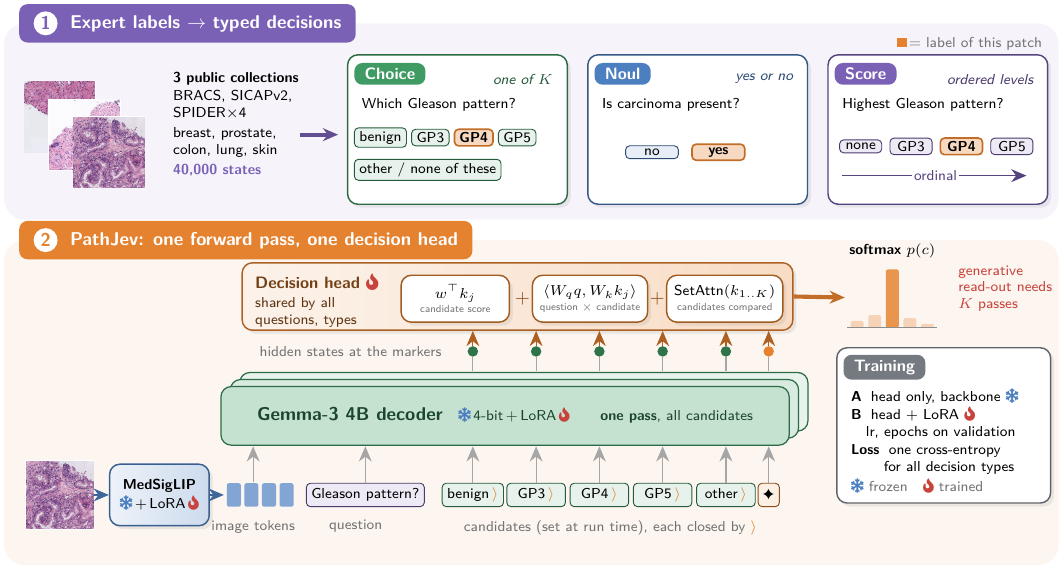}
\caption{Overview of PathJev. \textbf{(1)}~Expert labels of six training sets from three public collections are turned deterministically into typed decisions: a patch carries a Choice (one of $K$ candidates, including ``none of these''), Noul (yes/no) and Score (ordered levels) question; the orange candidate is the label. \textbf{(2)}~The image is encoded by MedSigLIP and enters the Gemma-3 decoder as tokens, followed by the question and all candidates, each closed by a marker token. One forward pass yields hidden states at the markers; a decision head shared by all questions sums a per-candidate score, a question--candidate interaction and a set-attention term, and a softmax returns the distribution. A generative read-out, as implemented here, needs one forward pass per candidate. Training: Stage~A fits the head on the frozen 4-bit backbone; Stage~B adds LoRA adapters. Whole-slide use is shown in Figure~\ref{fig:wsipipe}. Patches: SICAPv2 (CC BY 4.0). Bar heights in (2) are illustrative.}
\label{fig:overview}
\end{figure*}

In this section we describe PathJev. We first define the typed decisions it answers, then the model that turns an image and a run-time candidate set into a distribution over the candidates, and finally the training recipe; Figure~\ref{fig:overview} gives an overview.

\subsection{Typed decisions over runtime candidate sets}
We use the decision interface of System One models~\citep{almeida2026jev}, in the medical form of~\citet{tang2026omnimedjev}. A \emph{state} is an image (here a single histology patch) together with one or more \emph{decisions}. A decision $D=(Q,T,C)$ consists of a question $Q$, a type $T$ and a finite candidate set $C=\{c_1,\dots,c_K\}$ that is part of the model input rather than the index set of a fixed classifier. Three types are used: \textbf{Choice} (one of $K$ mutually exclusive candidates, e.g.\ the tissue class of a patch), \textbf{Noul} (a Bernoulli decision $\{\text{no},\text{yes}\}$, e.g.\ ``is tumour present?'') and \textbf{Score} (an ordered grid of levels, e.g.\ a Gleason grade group or a neoplasia level). For every decision the model returns a normalised distribution over its candidate set; one forward pass answers all decisions attached to a state.

\subsection{Model}
The backbone is MedGemma-1.5-4B~\citep{sellergren2026medgemmaonefive}, loaded in 4-bit NF4 precision. Candidate texts are written into the prompt after the image, each candidate closed by a key marker token and each question by a query marker token. One forward pass yields the last hidden states; a single decision head $f_\phi$, shared by all questions and decision types, reads the (layer-normalised) hidden state $k_{jk}$ at each candidate marker and $q_j$ at the question marker and returns one logit per candidate as the sum of three terms: a per-candidate linear score $w^\top k_{jk}$, a question--candidate interaction $\langle W_q q_j, W_k k_{jk}\rangle/\sqrt{d}$, and a set-attention term through which the candidates of one question attend to each other (multi-head attention with $d=128$, conditioned on $\log K$),
\begin{equation}
  z_{jk} = f_\phi\!\left(q_j,\, k_{j1},\dots,k_{jK}\right)_k, \qquad p_{jk} = \frac{\exp z_{jk}}{\sum_{l} \exp z_{jl}},
\end{equation}
so that no projection onto the language-model vocabulary is needed. All decision types share one objective, the cross-entropy between the target distribution $y_j$ induced by the annotation and the predicted $p_j$, averaged over the $M$ decisions in a batch: $\mathcal{L} = -\tfrac{1}{M}\sum_j \sum_k y_{jk}\log p_{jk}$. Choice and Noul targets are one-hot; Score targets are one-hot on the annotated level.

\subsection{Training recipe}
The decision head, the objective and the two-stage schedule follow~\citet{tang2026omnimedjev}; what is specific to PathJev-P is the data, described in Section~\ref{sec:data}. \textbf{Stage~A} trains the head alone for 200 optimiser steps with the backbone frozen. \textbf{Stage~B} trains the head together with LoRA adapters (rank 16, $\alpha=16$, dropout 0.05, all linear layers of the vision encoder and of the language model). Images are resized to $896\times896$ pixels; Choice candidates are shuffled at every presentation so that the model cannot rely on position, while Score candidates keep their order, which encodes the level. The effective batch is 32 states (8 per device, 4 accumulation steps). The learning rate and number of epochs are selected on validation data (Section~\ref{sec:setup}). We call the resulting patch-level model \textbf{PathJev-P}.

\paragraph{Input construction.} Every state of a source carries the same list of questions, in the same order and with the same candidate lists, whatever its class; the target is never part of the input, and the prompt header carries no dataset name. The text of a test prompt is therefore constant within a test set, and the model's answer to the main question can differ between images only through the image. The question wording does name the organ (e.g.\ ``this prostate tissue''), so the auxiliary organ question is answerable from the text; it is not an endpoint and is reported nowhere, but it enters the validation loss used for checkpoint selection. Selecting on the main Choice question alone would pick the same checkpoint in every run (\AuditSelRuns{} runs) and the same configuration for both methods.

\section{Data}
\label{sec:data}

In this section we describe the data used to train and test PathJev: how datasets are assigned to roles so that the test sets are free of documented overlap with the backbone's training data, how expert labels are turned into typed decisions, and which test sets probe in-domain performance and generalisation to unseen organs.

\subsection{Contamination-aware roles}
MedGemma's vision encoder was trained on data including TCGA and CAMELYON, and MedGemma~1.5 added further, partly unnamed, histopathology data~\citep{sellergren2026medgemmaonefive}; OmniMed-Jev saw PathMNIST, which derives from NCT-CRC-HE-100K~\citep{tang2026omnimedjev}. We therefore assign every dataset one role: \emph{train}, \emph{clean test} (no publicly documented overlap with the training data of the backbone, of OmniMed-Jev or of ours) or \emph{seen-source test} (same source as a training set, split by slide). ``Clean'' means \emph{no documented overlap}; undisclosed backbone data cannot be excluded, which is why the clean sets favour datasets released after or independently of the backbone. Every dataset, its licence and its role are listed in the data statement.

\subsection{Training and validation data}
PathJev-P is trained on patch states from six training sets drawn from three public collections: BRACS~\citep{brancati2022bracs} (breast lesion subtypes, official patient-level training split), SICAPv2~\citep{silvarodriguez2020sicapv2} (prostate Gleason patterns, patient-level split) and the four organ subsets of SPIDER~\citep{nechaev2025spider} (colorectal, thorax, breast, skin; re-split by slide). The mix contains \NTrainStates{} states, with BRACS and SICAPv2 in full and the four SPIDER subsets drawn class-stratified to equal size; validation uses \NValStates{} states drawn from the validation splits of the same sources. Each state carries (i) the source's main Choice question (lesion subtype, Gleason pattern or tissue class), with an ``other tissue / none of these'' candidate; (ii) an organ Choice over \NOrganCandidates{} organs shared by all sources; (iii) Noul presence questions and an ordinal Score question derived deterministically from the expert label (e.g.\ highest Gleason pattern, neoplasia level). All labels are expert annotations or deterministic functions of them; no LLM-generated labels are used.

\subsection{Test data}
\textbf{Clean, unseen domain.} HepatoBench~\citep{hepatobench2026} (liver, \NClassesHepato{} tissue classes) and HiESD~\citep{ge2025hiesd} (gastric endoscopic submucosal dissection, \NClassesHiesd{} classes; 104 slides from 44 patients, grouped by patient with the map released with the dataset). Neither organ has a positive training example; ``liver'' and ``stomach'' do appear as negative candidates of the training organ question. Several class concepts do occur in training (e.g.\ fibrosis, inflammation, necrosis and adenocarcinoma in SPIDER), so the unseen sets test new organs and partly new vocabularies, not entirely new concepts. The ``other'' candidate is never the target in training; at test time it is the target only for the residual-tissue class of HiESD. \textbf{Clean, in domain.} The official test splits of BRACS and SICAPv2, patient-disjoint from their training splits. \textbf{Seen source.} The SPIDER test slides of the four organs, reported separately and never in the headline: the SPIDER split is only slide-disjoint (no patient identifiers; the official split was not even slide-disjoint and was redrawn), and SPIDER derives from the HISTAI slide collection, which other pathology foundation models were trained on, so overlap with undisclosed backbone data is more plausible than for the other sets. Large test files are capped at \NTestCap{} states per set by a fixed, class-balanced draw shared by all methods (equal counts per class, limited by availability), so prevalence-dependent metrics (ECE, NLL, Brier) refer to balanced classes; the cap was set for compute reasons after the pilot and before any run reported here. On the full files, scored for every method except the generative baseline, no clean-set macro value changes by more than \FullMaxDiffBacc{} in balanced accuracy or \FullMaxDiffEce{} in ECE (Appendix~\ref{app:full}); the full files keep the original prevalences. Table~\ref{tab:testsets} lists the test sets. An audit of exact (decoded-pixel hash) and near (perceptual hash) duplicates across the training, validation and test images found no copy of a training or validation image in any clean test set and no copy across datasets; two SPIDER-skin test states duplicate training images from other slides of the same class and are excluded from the test set for every method (two more in the full file). The full HepatoBench release contains exact duplicate images, which do not enter the capped draw. The audit cannot detect patient-level links or heavily transformed copies.

\begin{table}[t]
\centering
\caption{Test sets of the main Choice question. States: fixed class-balanced draw shared by all methods (of the available states). Groups: units resampled by the cluster bootstrap. SPIDER has no patient identifiers, so its groups are finer than patients. $^\S$HepatoBench releases no patient or slide identifiers and consecutive image indices appear to come from the same tissue, so its images are resampled as independent units, which understates the uncertainty of comparisons that involve it.}
\label{tab:testsets}
\footnotesize
\IfFileExists{generated/tab_testsets.tex}{
\begin{tabular}{@{}lllrrrl@{}}
\toprule
Set & Organ & Role & Classes & States & Groups & Group unit \\
\midrule
HepatoBench & liver & clean, unseen domain & 7 & 5{,}000 of 87{,}301 & 5{,}000 & image$^\S$ \\
HiESD & stomach & clean, unseen domain & 10 & 5{,}000 of 23{,}679 & 44 & patient \\
BRACS & breast & clean, in domain & 7 & 570 & 30 & patient \\
SICAPv2 & prostate & clean, in domain & 4 & 2{,}122 & 21 & patient \\
SPIDER colorectal & colon & seen source & 13 & 5{,}000 of 7{,}485 & 158 & slide \\
SPIDER thorax & lung & seen source & 13 & 5{,}000 of 5{,}569 & 38 & slide \\
SPIDER breast & breast & seen source & 18 & 5{,}000 of 10{,}063 & 76 & slide \\
SPIDER skin & skin & seen source & 24 & 4{,}998 of 15{,}485 & 309 & slide \\
\bottomrule
\end{tabular}
}{\textcolor{red}{[table pending]}}
\end{table}

\section{Experiments}
\label{sec:results}

In this section we first describe the compared methods, model selection and evaluation, then compare PathJev-P with the baselines, analyse its calibration, including how the comparison with the generative model depends on the way that model's probabilities are read out, and finally illustrate its use on whole slides.

\subsection{Setup}
\label{sec:setup}
\paragraph{Compared methods.} All methods answer the same questions about the same images, with the same candidate sets and candidate texts.

\textbf{(a) OmniMed-Jev}~\citep{tang2026omnimedjev}, the released decision-native checkpoint, used zero-shot. It tells us how far a general medical decision model already goes on histopathology; since it differs from PathJev-P in training data, training length, model selection and input pre-processing, the comparison is between two systems.

\textbf{(b) Generative SFT.} The same backbone and LoRA configuration, fine-tuned on the same training states to generate the answer text, with every decision of a state turned into one training example that lists the candidates as options. Its probability for a candidate is the softmax, over the candidate set, of the summed log-probability of the candidate's tokens given the image and the prompt, the read-out used by OmniMed-Jev; one forward pass is needed per candidate. Both trained methods search the same grid and use the same selection rule, but they differ in objective, in the number of supervised targets and in optimiser steps (Appendix~\ref{app:training}), so the comparison is between two systems rather than an ablation of the decision head.

\textbf{(c) CONCH}~\citep{lu2024conch}, a contrastive pathology vision--language model, used zero-shot with an ensemble of five prompt templates per candidate.

\textbf{(d) Linear probe} on H-optimus-0 features~\citep{saillard2024hoptimuszero}, trained on the labels of the training sets. Where a test question has no training labels (HepatoBench, HiESD), the probe is fitted on part of the test set itself by nested, grouped cross-validation. It is therefore a supervised reference, not a competitor.

\textbf{(e) Inner-Jev read-out}~\citep{geuter2026innerjev}: the base MedGemma-1.5-4B, zero-shot, with the candidates listed as lettered options and the distribution read from the next-token probabilities of the letters.

Implementation details of every method are in Appendix~\ref{app:details}.

\paragraph{Model selection.} For PathJev-P and the generative baseline we search learning rates $\{1,2\}\times10^{-4}$ and 3 or 5 epochs with one seed, keep the checkpoint and configuration with the lowest validation negative log-likelihood (NLL) on the Choice questions, and train the selected configuration with two further seeds. Only the validation mix (\NValStates{} states) is used; no test data enters any choice of checkpoint, configuration, prompt or candidate wording.

\paragraph{Evaluation.} We evaluate the main Choice question of each test set (tissue class, lesion subtype or Gleason pattern) with balanced accuracy and expected calibration error (ECE; 15 equal-width bins on the top-candidate confidence), averaged over the clean test sets and reported separately for the unseen organs and the in-domain sets. Negative log-likelihood, Brier score, macro-F1, adaptive ECE, the area under the risk--coverage curve and reliability diagrams~\citep{guo2017calibration,nixon2019measuring} complement them. Confidence intervals are 95\% intervals of a paired cluster bootstrap that resamples patients (slides or images where patients are not identified) within each test set, and the trained methods are compared seed by seed. The hypotheses, endpoints, test sets and tests were fixed in an analysis plan before any test-set evaluation; analyses added later are labelled post hoc. Appendix~\ref{app:protocol} gives the statistical procedure, the plan and its amendments.

\begin{table*}[t]
\centering
\caption{Balanced accuracy (bAcc) and expected calibration error (ECE) on the main Choice question, averaged over the sets of each group; the clean average covers the unseen organs (HepatoBench, HiESD) and the in-domain sets (BRACS, SICAPv2). Trained methods: mean\,$\pm$\,SD over \NSeeds{} seeds. Bold: best language-model-based value per column. $^\dagger$On the unseen organs the probe is fitted with labels of the test domain.}
\label{tab:main}
\footnotesize
\setlength{\tabcolsep}{1.6pt}
\IfFileExists{generated/tab_main.tex}{\begin{tabular}{@{}lcccccccc@{}}
\toprule
 & \multicolumn{2}{c}{Clean average} & \multicolumn{2}{c}{Unseen organs} & \multicolumn{2}{c}{In domain} & \multicolumn{2}{c}{Seen source} \\
\cmidrule(lr){2-3} \cmidrule(lr){4-5} \cmidrule(lr){6-7} \cmidrule(lr){8-9}
Method & bAcc$\uparrow$ & ECE$\downarrow$ & bAcc$\uparrow$ & ECE$\downarrow$ & bAcc$\uparrow$ & ECE$\downarrow$ & bAcc$\uparrow$ & ECE$\downarrow$ \\
\midrule
\multicolumn{9}{@{}l}{\textit{Trained on our data}} \\
\hspace{0.6em}PathJev-P (ours) & \textbf{0.541$_{\pm.015}$} & \textbf{0.204$_{\pm.002}$} & \textbf{0.390$_{\pm.015}$} & \textbf{0.301$_{\pm.012}$} & 0.692$_{\pm.019}$ & \textbf{0.108$_{\pm.013}$} & 0.781$_{\pm.017}$ & 0.049$_{\pm.025}$ \\
\hspace{0.6em}(b) Gen-SFT & 0.492$_{\pm.006}$ & 0.279$_{\pm.006}$ & 0.277$_{\pm.026}$ & 0.446$_{\pm.017}$ & \textbf{0.707$_{\pm.015}$} & 0.113$_{\pm.018}$ & \textbf{0.810$_{\pm.006}$} & \textbf{0.044$_{\pm.011}$} \\
\addlinespace[3pt]
\multicolumn{9}{@{}l}{\textit{Zero-shot}} \\
\hspace{0.6em}(a) OmniMed-Jev & 0.213 & 0.517 & 0.253 & 0.540 & 0.173 & 0.493 & 0.212 & 0.577 \\
\hspace{1.4em}without ``other'' & 0.311 & 0.322 & &  & &  & &  \\
\hspace{1.4em}permuted & 0.217 & 0.521 & &  & &  & &  \\
\hspace{0.6em}(c) CONCH & 0.345 & 0.403 & 0.328 & 0.487 & 0.363 & 0.319 & 0.432 & 0.394 \\
\hspace{0.6em}(e) Inner-Jev & 0.288 & 0.578 & 0.230 & 0.636 & 0.346 & 0.521 & 0.142 & 0.688 \\
\addlinespace[3pt]
\multicolumn{9}{@{}l}{\textit{Supervised reference}} \\
\hspace{0.6em}(d) Linear probe$^\dagger$ & 0.724 & 0.047 & 0.772 & 0.021 & 0.677 & 0.074 & 0.905 & 0.013 \\
\bottomrule
\end{tabular}
}{\textcolor{red}{[table generated from the final analysis]}}
\end{table*}

\subsection{Comparison with the baselines}
On the clean test sets PathJev-P reaches a balanced accuracy of \ResPJCleanBacc{}\,$\pm$\,\ResPJCleanBaccSD{} (Table~\ref{tab:main}), against \ResSFTCleanBacc{} for the generative baseline, \ResCONCHCleanBacc{} for CONCH, \ResIJCleanBacc{} for the Inner-Jev read-out and \ResOMJCleanBacc{} for the released OmniMed-Jev. The paired difference is \HOneBDiffSFT{} [\HOneBLoSFT{}, \HOneBHiSFT{}] against the generative baseline and \HOneBDiffCONCH{} [\HOneBLoCONCH{}, \HOneBHiCONCH{}] against CONCH, and it favours PathJev-P in each of the three seeds. Its probabilities are also better calibrated than those of the four comparators (a), (b), (c) and (e), with the read-outs fixed in the analysis plan: the ECE is \ResPJCleanEce{}\,$\pm$\,\ResPJCleanEceSD{} against \ResSFTCleanEce{} for the generative baseline (difference \HOneADiffSFT{} [\HOneALoSFT{}, \HOneAHiSFT{}]) and \ResCONCHCleanEce{} for CONCH (\HOneADiffCONCH{} [\HOneALoCONCH{}, \HOneAHiCONCH{}]). These were the two hypotheses of the plan. The accuracy result is robust; the calibration result against the generative baseline depends on how that model's probabilities are read out (Section~\ref{sec:robust}).

The released OmniMed-Jev often answers ``other''. Removing that candidate raises its balanced accuracy, but it stays far below PathJev-P (Table~\ref{tab:main}), and its weakness is not a matter of input size: with images pre-resized to 896 instead of 448 pixels it reaches \AThreeOMJNineSixCleanBacc{}.

On the two organs absent from training, PathJev-P reaches a balanced accuracy of \ResPJUnseenBacc{}, against \ResSFTUnseenBacc{} for the generative baseline and \ResCONCHUnseenBacc{} for CONCH. On the in-domain sets the two trained methods are close (\ResPJInDomBacc{} and \ResSFTInDomBacc{}), and on the seen-source SPIDER sets the generative baseline is slightly ahead (\ResSFTSeenBacc{} against \ResPJSeenBacc{}). Set by set (Appendix~\ref{app:perset}), PathJev-P is more accurate than the generative baseline on both unseen organs, whereas its advantage over CONCH comes from HepatoBench; on HiESD, CONCH is slightly more accurate. The secondary metrics and the Noul and Score questions show the same pattern (Appendix~\ref{app:secondary}). The comparisons keep their sign when HepatoBench, whose images cannot be grouped by slide, is removed: the difference in balanced accuracy against the generative baseline is then \AThreeHepNoHepHOneBDiffSFT{} [\AThreeHepNoHepHOneBLoSFT{}, \AThreeHepNoHepHOneBHiSFT{}] (Appendix~\ref{app:a3}).

Even so, accuracy on new organs remains modest, and on HiESD, with ten gastric classes, it is low for every language-model-based method (Appendix~\ref{app:perset}).

Two post-hoc checks (Appendix~\ref{app:a3}) support this comparison. The order in which the candidates are listed matters little for PathJev-P: over \AThreePermPJN{} random orders, its clean balanced accuracy stays between \AThreePermPJBaccMin{} and \AThreePermPJBaccMax{}, although \AThreePermPJFlip{}\% of individual predictions change. The generative baseline behaves similarly, while OmniMed-Jev changes \AThreePermOMJFlip{}\% of its predictions. Finally, with every image replaced by a uniform grey image, PathJev-P and the generative baseline give one prediction per test set and their balanced accuracy falls to chance, so the question text, which is the same for every image of a set, does not by itself separate the classes.

The linear probe on H-optimus-0 features answers a different question: how well a classifier does when it is trained on labels of the task. On the in-domain sets, where it is trained on the same labels as PathJev-P, the two are comparable in balanced accuracy (\ResProbeInDomBacc{} for the probe) but the probe is better calibrated (ECE \ResProbeInDomEce{}). On the seen-source sets it is far ahead (\ResProbeSeenBacc{}), and on the unseen organs, where it is fitted on labelled patches of the test domain itself, it reaches \ResProbeUnseenBacc{}. Where labels for a fixed question exist, a probe remains the stronger tool.

\subsection{Calibration}
\label{sec:robust}
The generative baseline's probabilities are reconstructed from the log-likelihoods of the candidate texts, and the way this is done is a choice. In a post-hoc check (Appendix~\ref{app:a3}) we therefore repeated the comparison with two other read-outs.

Dividing each candidate's summed log-likelihood by its number of tokens leaves the generative baseline's balanced accuracy almost unchanged (\AThreeSFTLenCleanBacc{} on the clean sets, \AThreeSFTLenUnseenBacc{} on the unseen organs) but changes its calibration substantially (Table~\ref{tab:a3sftvar}). Its clean ECE becomes \AThreeSFTLenCleanEce{}, similar to PathJev-P's \ResPJCleanEce{}, with a lower NLL (\AThreeSFTLenCleanNll{} against \ResPJCleanNll{}); on the unseen organs it becomes better calibrated than PathJev-P (ECE \AThreeSFTLenUnseenEce{} against \ResPJUnseenEce{}), and in domain worse (\AThreeSFTLenInDomEce{} against \ResPJInDomEce{}). Adding the probability of the end-of-answer token changes nothing. The calibration advantage of PathJev-P over the generative model therefore holds for the read-out used by OmniMed-Jev and fixed in our plan, but not in general; its accuracy advantage holds for every read-out.

\begin{figure}[t]
\centering
\includegraphics[width=\linewidth]{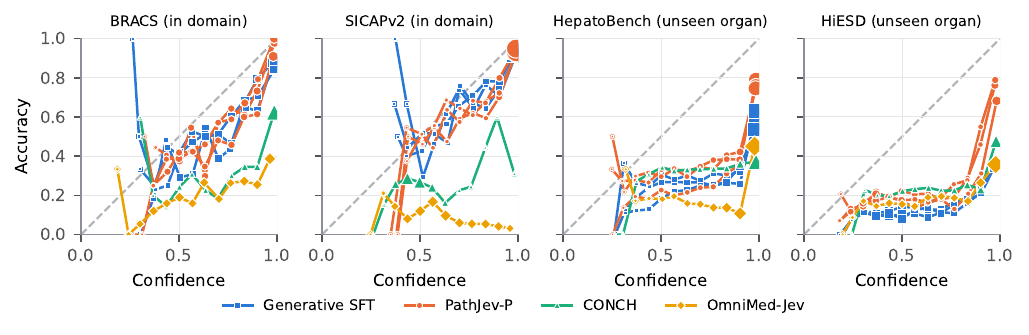}
\caption{Reliability diagrams of the main Choice question before temperature scaling, one panel per clean test set: accuracy against top-candidate confidence in 15 equal-width bins (every non-empty bin; marker area $\propto$ share of the set's states). Trained methods: one line per seed. Sparse bins (small markers) are noisy. The dashed diagonal is perfect calibration.}
\label{fig:reliability}
\end{figure}

A single temperature fitted on the validation mix lowers the clean-set ECE of every method (per set in Table~\ref{tab:scaled}; PathJev-P \ResPJCleanEce{}$\to$\ResPJCleanEceT{}, generative baseline \ResSFTCleanEce{}$\to$\ResSFTCleanEceT{}). For the zero-shot models the fitted temperatures are large and flatten their outputs, which lowers their ECE (CONCH \ResCONCHCleanEce{}$\to$\ResCONCHCleanEceT{}) without making them more discriminative. Measured by NLL, which rewards both, PathJev-P is the best language-model-based method on the clean sets after scaling (\ResPJCleanNllT{} against \ResCONCHCleanNllT{} for CONCH and \ResSFTCleanNllT{} for the generative baseline). On the unseen organs, however, scaled CONCH reaches a lower NLL (\TmpCONCHValNll{} against \TmpPJValNll{}): the reliability diagrams (Figure~\ref{fig:reliability}) show that every model is over-confident on these organs, and a temperature fitted on the training domains does not correct this for PathJev-P.

In a post-hoc analysis we fitted the temperature on different data (Table~\ref{tab:temperature}). A temperature fitted on one unseen organ and applied to the other lowers PathJev-P's NLL on the unseen organs from \TmpPJValNll{} to \TmpPJCrossNll{}, close to the \TmpPJOracleNll{} of an oracle fitted on the evaluated set itself. With two organs this is an observation, not a transfer property, but it suggests that labelled data from one new organ can correct much of the over-confidence on another.

\begin{table}[t]
\centering
\caption{Post hoc: negative log-likelihood on the unseen organs (lower is better) after one temperature per method, fitted on different data; seed mean for trained methods. Bold: best language-model-based value per column. The oracle is fitted on the evaluated set itself and is not a deployable procedure. Brier scores in Table~\ref{tab:temperaturebrier}.}
\label{tab:temperature}
\footnotesize
\IfFileExists{generated/tab_temperature.tex}{\begin{tabular}{@{}lccccc@{}}
\toprule
& \multicolumn{5}{c}{Temperature fitted on} \\
\cmidrule(l){2-6}
Method & none & val.\ mix & in-domain val. & other unseen organ & same set (oracle) \\
\midrule
PathJev-P (ours) & \textbf{2.39} & 2.08 & \textbf{1.99} & 1.78 & \textbf{1.71} \\
(b) Gen-SFT & 3.70 & 2.87 & 2.66 & 2.03 & 1.92 \\
(a) OmniMed-Jev & 3.76 & 1.98 & 2.09 & 1.97 & 1.97 \\
(c) CONCH & 4.41 & \textbf{1.77} & 2.22 & \textbf{1.77} & 1.75 \\
(e) Inner-Jev & 5.06 & 2.15 & 2.18 & 2.17 & 2.11 \\
\midrule
(d) Linear probe$^\dagger$ & 0.64 & 0.64 & 0.64 & 0.64 & 0.64 \\
\bottomrule
\end{tabular}
}{\textcolor{red}{[table generated from the final analysis]}}
\end{table}

\subsection{Whole-slide demonstration}
\begin{figure*}[t]
\centering
\includegraphics[width=\textwidth]{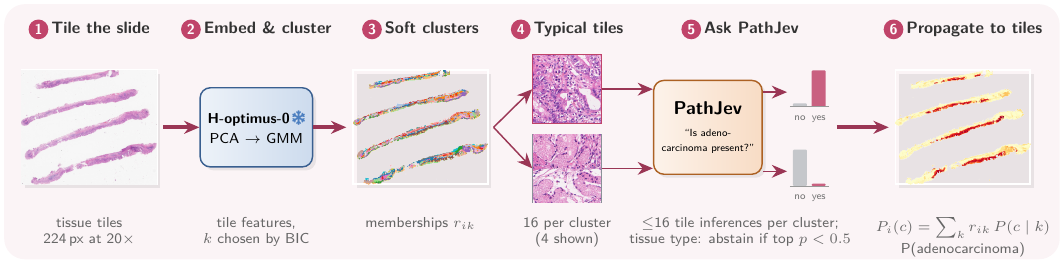}
\caption{Whole-slide use of PathJev. Tiles of the slide are embedded with H-optimus-0 and clustered (PCA, Gaussian mixture with the number of components chosen by BIC, soft memberships $r_{ik}$). PathJev-P is run on the 16 most typical tiles of each cluster (4 shown for two clusters; one forward pass per tile answers all questions), and the cluster's distribution $P(c\mid k)$ is the mean of their raw probabilities. A cluster abstains when the top probability of its tissue-type distribution (11 candidates) is below 0.5. Cluster probabilities are propagated to every tile as $P_i(c)=\sum_k r_{ik}P(c\mid k)$; their calibration was not evaluated. HiESD slide of Figure~\ref{fig:wsi} (CC BY 4.0); bar heights are illustrative.}
\label{fig:wsipipe}
\end{figure*}
\begin{figure*}[t]
\centering
\includegraphics[width=0.8\textwidth]{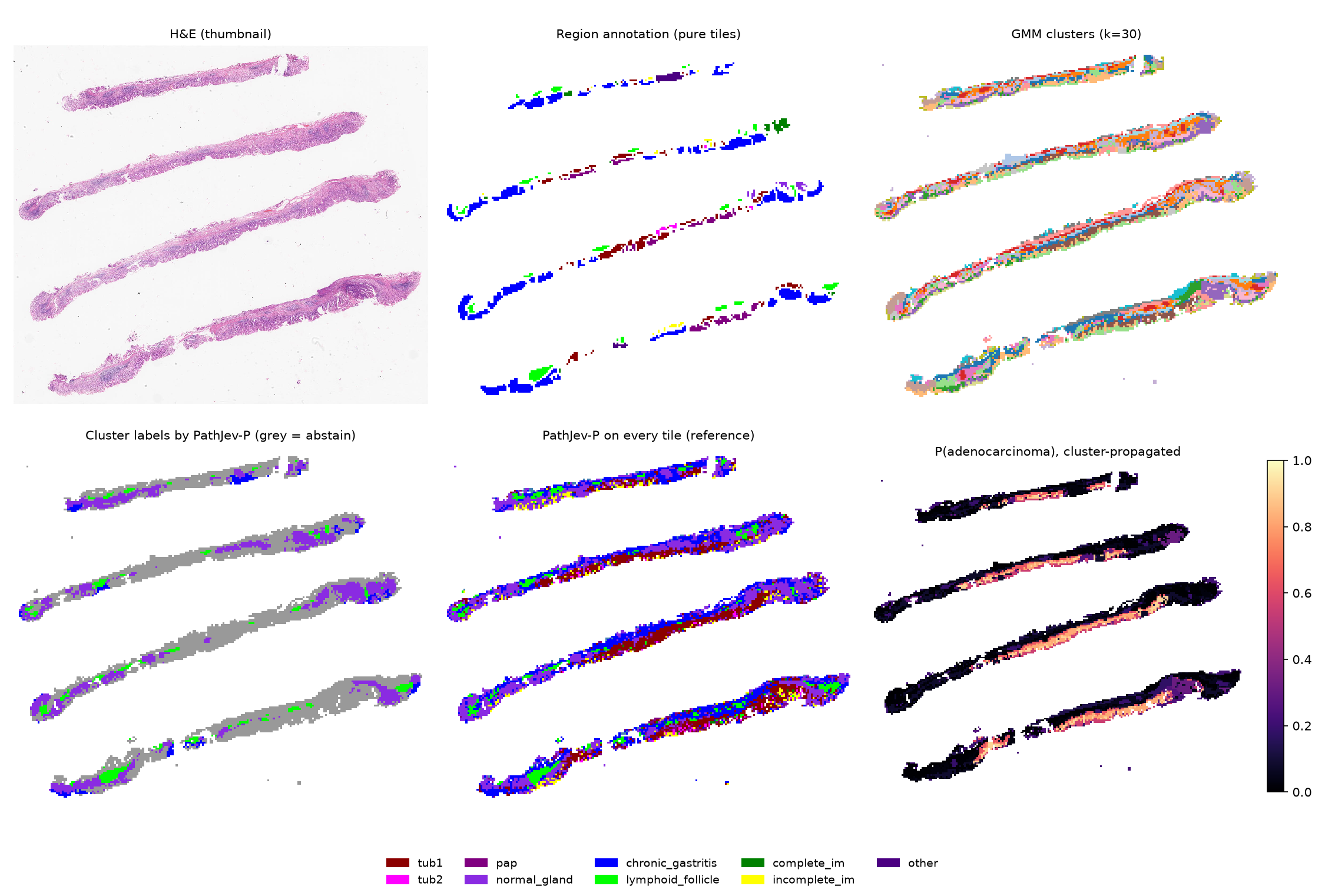}\\[2pt]
\includegraphics[width=0.8\textwidth]{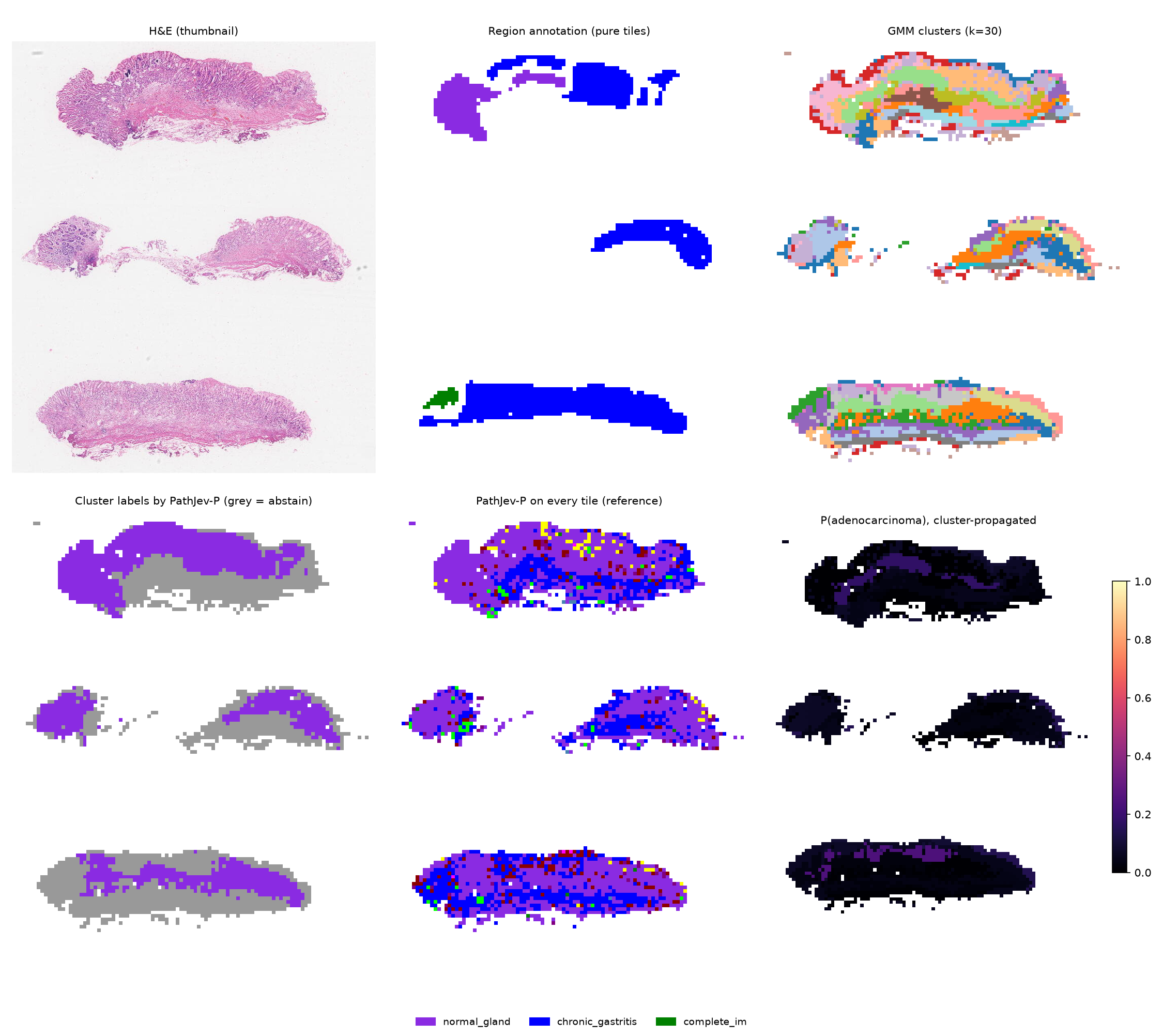}
\caption{Whole-slide demonstration on two of the four HiESD slides (top: slide with adenocarcinoma; bottom: slide without; the other two are in Appendix~\ref{app:wsi}). Panels: thumbnail; region annotation (tiles whose mask is $\geq$90\% one class); GMM clusters of H-optimus-0 tile features; cluster labels from PathJev-P (grey: cluster abstains, top probability $<0.5$); PathJev-P applied to every tile (reference); probability of adenocarcinoma propagated from the clusters. The number of clusters chosen by BIC reached the upper bound of the search range (30) on three of the four slides.}
\label{fig:wsi}
\end{figure*}
To illustrate the intended use (Figure~\ref{fig:wsipipe}), we cluster the tiles of four HiESD whole slides (stomach, unseen in training; drawn at random with a fixed seed among the slides whose annotation has at least three classes; they come from three patients, and the first slide inspected and one of the slides without carcinoma belong to the same patient) and let PathJev-P label each cluster from its most typical tiles. Pipeline parameters, including the abstention threshold, were fixed before the first run; PCA and the mixture model are fitted on each slide without labels. The fine-grained gastric categories are mostly not recognised and most clusters abstain. The adenocarcinoma presence question separates carcinoma from non-carcinoma tiles of the region annotation on the two slides that contain carcinoma, and few tiles of the two slides without carcinoma exceed $P=0.5$ (Table~\ref{tab:wsi}). These metrics were added after inspecting the first slide, which is one of the two with carcinoma; they are computed only on tiles whose annotation is at least 90\% one class, which excludes boundary and mixed regions; and four slides support no claim about diagnostic sensitivity or about the abstention. The clusters require PathJev-P on far fewer tiles than labelling every tile, but every tile is still embedded with H-optimus-0, and we did not measure end-to-end time.

\begin{table}[t]
\centering
\caption{Whole-slide demonstration, adenocarcinoma presence question against the region annotation (descriptive; metrics added after inspecting slide 1a7624bc). Labelled: tiles whose annotation is at least 90\% one class, the only tiles evaluated. PathJev-P tiles: cluster representatives run through the model. Sensitivity and specificity at $P(\text{yes})>0.5$; --: no carcinoma in the slide.}
\label{tab:wsi}
\scriptsize
\setlength{\tabcolsep}{3.5pt}
\IfFileExists{generated/tab_wsi.tex}{\begin{tabular}{@{}llrrrrcccccc@{}}
\toprule
& & & & & PathJev-P & \multicolumn{2}{c}{AUROC} & \multicolumn{2}{c}{Sensitivity} & \multicolumn{2}{c}{Specificity} \\
\cmidrule(lr){7-8} \cmidrule(lr){9-10} \cmidrule(l){11-12}
Slide & Patient & Tiles & Labelled & Clusters & tiles & clusters & all tiles & clusters & all tiles & clusters & all tiles \\
\midrule
1a7624bc & P22 & 2{,}731 & 713 & 30 & 471 & 0.983 & 0.946 & 0.912 & 0.912 & 0.993 & 0.845 \\
46e7dccd & P27 & 7{,}165 & 2{,}073 & 30 & 480 & 0.942 & 0.928 & 0.920 & 0.824 & 0.896 & 0.879 \\
c6a80707 & P38 & 3{,}279 & 1{,}388 & 30 & 480 & -- & -- & -- & -- & 1.000 & 0.970 \\
e492f29a & P22 & 1{,}043 & 230 & 18 & 288 & -- & -- & -- & -- & 1.000 & 0.987 \\
\bottomrule
\end{tabular}
}{\textcolor{red}{[table pending]}}
\end{table}

\section{Discussion and limitations}
\label{sec:discussion}

In this section we discuss what the results show, when a decision-native model is the right tool, and the limitations of this study.

\paragraph{What the comparison shows.} Trained on the same states with the same backbone, PathJev-P is more accurate than a generative model that scores the candidate texts, and the gain lies on the organs absent from training. The two systems differ in more than the decision head (objective, number of supervised targets, optimiser steps), so the gain cannot be attributed to the head alone; one explanation, which we have not tested, is that the head is trained directly on distributions over candidate sets, whereas the generative distribution is only assembled at test time from text log-likelihoods. The calibration comparison is less clear-cut: PathJev-P is better calibrated than the generative model under the summed read-out used in previous work, but not under a length-normalised one. Calibration advantages of decision-native models over generative baselines should therefore be reported for more than one read-out. A practical difference remains in any case: PathJev-P answers in one forward pass, the generative read-out in one pass per candidate.

\paragraph{When to use it.} On new organs PathJev-P is over-confident, and a temperature fitted on the training domains does not correct this; in our post-hoc analysis, labelled data from one new organ corrected most of it on the other, which suggests a cheap remedy that remains to be tested on more organs. Where labels for a fixed question exist, a linear probe on a pathology foundation model is the better tool: it is better calibrated in every setting and far more accurate with target-domain labels. The case for a decision-native model is the open-vocabulary, label-free setting, such as naming tissue clusters in whole slides, where our demonstration shows that cluster-level labelling with abstention is feasible for coarse questions; its accuracy on fine-grained classes of new organs (\ResPJUnseenBacc{} balanced accuracy) is not yet sufficient.

\paragraph{Limitations.} All states are single patches, and the whole-slide pipeline was not evaluated systematically against pathologists. ``Clean'' means no documented overlap with the backbone's training data, which is partly undisclosed. Only two organs are unseen, HepatoBench images cannot be grouped by slide or patient, and SPIDER only by slide. We used one backbone in 4-bit precision, one wording per candidate and one prompt ensemble for the zero-shot models, and the generative baseline was scored only on the capped test sets. Several training datasets are licensed for non-commercial use only. Training on mosaics of cluster tiles and validating the whole-slide pipeline with pathologists are the next steps.

\section*{Acknowledgements}
We acknowledge the use of the ADHOC (Astrophysical Data HPC Operating Center) resources, within the project ``Strengthening the Italian Leadership in ELT and SKA (STILES)'', proposal nr.\ IR0000034, admitted and eligible for funding from the funds referred to in the D.D.\ prot.\ no.\ 245 of August 10, 2022 and D.D.\ 326 of August 30, 2022, funded under the program ``Next Generation EU'' of the European Union, ``Piano Nazionale di Ripresa e Resilienza'' (PNRR) of the Italian Ministry of University and Research (MUR), ``Fund for the creation of an integrated system of research and innovation infrastructures'', Action 3.1.1 ``Creation of new IR or strengthening of existing IR involved in the Horizon Europe Scientific Excellence objectives and the establishment of networks''.

\section*{Data statement}
All datasets are public and were used under their published licences, for research only. We do not redistribute the datasets; the figures reproduce a few example patches and slide maps from SICAPv2 and HiESD (both CC BY 4.0), with attribution. Labels are the expert annotations released with each dataset or deterministic functions of them. SPIDER is derived from HISTAI slides and is therefore reported only as a seen-source test. None of the datasets is documented to contain TCGA, CAMELYON or NCT-CRC-HE material, the known histopathology sources of the backbone and of OmniMed-Jev; undisclosed backbone training data cannot be excluded. Non-commercial licences are flagged.
\begin{center}
\scriptsize
\IfFileExists{generated/data_statement.tex}{
\begin{tabular}{@{}>{\raggedright}p{2.5cm}>{\raggedright}p{3.4cm}>{\raggedright}p{2.2cm}>{\raggedright}p{2.3cm}>{\raggedright\arraybackslash}p{3.7cm}@{}}
\toprule
Dataset & DOI / ID & Licence & Access & Role in this work \\
\midrule
SPIDER (4 organs) & \nolinkurl{arXiv:2503.02876} & CC BY-NC 4.0 & gated (accept conditions) & train + seen-source test (re-split by slide; SPIDER is a HISTAI subset) \\
BRACS & \nolinkurl{10.1093/database/baac093} & CC BY-NC 4.0$^\ddagger$ & registration, terms of use & train + clean in-domain test (official patient-level split) \\
SICAPv2 & \nolinkurl{10.17632/9xxm58dvs3.2} & CC BY 4.0 & open download & train + clean in-domain test (official patient-level split) \\
HepatoBench & \nolinkurl{10.57967/hf/8231} & CC BY-NC 4.0$^\ddagger$ & open download & clean test, unseen domain (whole dataset held out) \\
HiESD & \nolinkurl{10.6084/m9.figshare.28919840.v1} & CC BY 4.0 & open download & clean test, unseen domain, and WSI demonstration (whole dataset held out) \\
\bottomrule
\end{tabular}
}{\textcolor{red}{[table pending]}}
\end{center}
{\scriptsize $^\ddagger$The dataset's own pages state different licences; the most restrictive one is recorded.\par}

\bibliographystyle{unsrtnat}
\bibliography{references}

\appendix
\section{Analysis protocol}
\label{app:protocol}
This appendix details the statistical procedure, the model-selection choices and the history of the analysis plan.

\paragraph{Statistics.} Each test set is resampled by cluster: its groups (patients for BRACS, SICAPv2 and HiESD, slides for SPIDER, images for HepatoBench; Table~\ref{tab:testsets}) are drawn with replacement, every metric is recomputed on the replicate (balanced accuracy over the classes present in it) and the macro average is recomputed; the same \NResamples{} replicates are used for every method and seed. We report marginal 95\% percentile intervals of the paired differences. The two hypotheses of the plan were that PathJev-P has a lower ECE (H1a) and a higher balanced accuracy (H1b) than methods (a), (b), (c) and (e) on the clean test sets. They were tested with one-sided bootstrap $p$-values, Holm-adjusted within each family of four comparisons (ECE, and balanced accuracy with a non-inferiority margin of $-0.02$ against the generative baseline and $0$ against the others, each against methods (a), (b), (c) and (e)); every $p$-value reached the smallest attainable value, $1/(\NResamples{}+1)$. Metrics of trained methods are averaged over seeds (probabilities are never averaged), and a claim required the seed-mean effect and at least two of three seeds, each seed of PathJev-P paired with the same seed of the generative baseline. The intervals are conditional on the trained models and do not include the variance of retraining.

\paragraph{Selection details.} The grid is run with seed 1, and that run is also the first of the three final seeds. Selecting on the main Choice question alone, without the organ question, would pick the same checkpoint in every run (\AuditSelRuns{} runs) and the same configuration for both methods. The validation mix is also used to fit the temperature of Section~\ref{sec:results}, so the temperature is fitted on data already used for selection.

\paragraph{Plan and amendments.} The analysis plan, with hypotheses, endpoints, test sets, statistical tests and the decision rule for the next stage of the project, was fixed and hashed on 17~October~2026, before any test-set evaluation. Three amendments were appended. A1 capped the large test sets for compute reasons after a pilot (one seed, one epoch) that had been evaluated on class-stratified subsets of the same test files; no model or analysis choice was based on the pilot's test results. A2 (the temperature analyses of Section~\ref{sec:results}) was specified after interim results had been seen, and A3 (Section~\ref{sec:robust} and Appendix~\ref{app:a3}) after the full results and an external review; both are reported as post hoc. The plan is released with the code (\texttt{docs/stage1\_analysis\_plan.md}); the SHA-256 hashes of its versions begin with \texttt{68a18d157e6058d1} (original), \texttt{c21d0e5d279cb0ac} (A1), \texttt{bfe701859e136f90} (A2) and \texttt{497508189b8b6971} (A3), and \texttt{docs/stage1\_plan\_versions.md} gives the full hashes and the rule that rebuilds each version.

\section{Secondary analysis on the full test files}
\label{app:full}
This appendix repeats the main analysis on the full test files instead of the capped draws, as specified in the analysis plan.

\begin{table}[h]
\centering
\caption{Secondary analysis (plan amendment A1): the main metrics on the full test files instead of the capped draws, for every method except the generative baseline (not scored on the full files for compute reasons); seed mean for PathJev-P. BRACS and SICAPv2 are not capped, so the in-domain columns equal Table~\ref{tab:main}.}
\label{tab:full}
\small
\IfFileExists{generated/tab_full.tex}{\begin{tabular}{@{}lcccccc@{}}
\toprule
 & \multicolumn{2}{c}{Clean average} & \multicolumn{2}{c}{Unseen organs} & \multicolumn{2}{c}{In domain} \\
\cmidrule(lr){2-3} \cmidrule(lr){4-5} \cmidrule(lr){6-7}
Method & bAcc$\uparrow$ & ECE$\downarrow$ & bAcc$\uparrow$ & ECE$\downarrow$ & bAcc$\uparrow$ & ECE$\downarrow$ \\
\midrule
PathJev-P (ours) & \textbf{0.542} & \textbf{0.199} & \textbf{0.392} & \textbf{0.290} & \textbf{0.692} & \textbf{0.108} \\
(a) OmniMed-Jev & 0.213 & 0.491 & 0.254 & 0.489 & 0.173 & 0.493 \\
(c) CONCH & 0.349 & 0.415 & 0.335 & 0.510 & 0.363 & 0.319 \\
(e) Inner-Jev & 0.288 & 0.609 & 0.230 & 0.696 & 0.346 & 0.521 \\
\midrule
(d) Linear probe$^\dagger$ & 0.718 & 0.049 & 0.760 & 0.024 & 0.677 & 0.074 \\
\bottomrule
\end{tabular}
}{\textcolor{red}{[table pending]}}
\end{table}

\section{Results per test set}
\label{app:perset}
This appendix reports the main Choice question for each test set separately, without averaging over sets.

\begin{table}[h]
\centering
\caption{Main Choice question per test set (no averaging over sets): balanced accuracy, ECE and NLL before temperature scaling; seed mean for trained methods. The probe is cross-fitted within HepatoBench and HiESD.}
\label{tab:perset}
\scriptsize
\setlength{\tabcolsep}{3pt}
\IfFileExists{generated/tab_perset.tex}{\begin{tabular}{@{}llcccccc@{}}
\toprule
Set & Metric & PathJev-P & Gen-SFT & OmniMed-Jev & CONCH & Inner-Jev & Probe$^\dagger$ \\
\midrule
\multicolumn{8}{@{}l}{\textit{Unseen organs}} \\
HepatoBench & bAcc$\uparrow$ & \textbf{0.512} & 0.414 & 0.267 & 0.354 & 0.303 & 0.971 \\
 & ECE$\downarrow$ & \textbf{0.305} & 0.426 & 0.575 & 0.496 & 0.525 & 0.007 \\
 & NLL$\downarrow$ & \textbf{1.862} & 3.240 & 3.798 & 3.911 & 3.671 & 0.098 \\
\addlinespace[2pt]
HiESD & bAcc$\uparrow$ & 0.267 & 0.140 & 0.239 & \textbf{0.301} & 0.158 & 0.572 \\
 & ECE$\downarrow$ & \textbf{0.297} & 0.465 & 0.506 & 0.478 & 0.746 & 0.034 \\
 & NLL$\downarrow$ & \textbf{2.926} & 4.162 & 3.731 & 4.904 & 6.450 & 1.180 \\
\midrule
\multicolumn{8}{@{}l}{\textit{In domain}} \\
BRACS & bAcc$\uparrow$ & 0.582 & \textbf{0.598} & 0.214 & 0.440 & 0.266 & 0.596 \\
 & ECE$\downarrow$ & \textbf{0.141} & 0.151 & 0.428 & 0.393 & 0.585 & 0.113 \\
 & NLL$\downarrow$ & \textbf{1.092} & 1.127 & 2.743 & 3.089 & 3.981 & 1.057 \\
\addlinespace[2pt]
SICAPv2 & bAcc$\uparrow$ & 0.803 & \textbf{0.816} & 0.131 & 0.285 & 0.425 & 0.757 \\
 & ECE$\downarrow$ & \textbf{0.075} & \textbf{0.075} & 0.557 & 0.244 & 0.457 & 0.034 \\
 & NLL$\downarrow$ & \textbf{0.462} & 0.499 & 2.806 & 1.953 & 2.749 & 0.506 \\
\midrule
\multicolumn{8}{@{}l}{\textit{Seen source}} \\
SPIDER colorectal & bAcc$\uparrow$ & 0.831 & \textbf{0.848} & 0.299 & 0.359 & 0.229 & 0.903 \\
 & ECE$\downarrow$ & \textbf{0.021} & 0.036 & 0.524 & 0.475 & 0.637 & 0.013 \\
 & NLL$\downarrow$ & 0.422 & \textbf{0.394} & 4.224 & 5.167 & 6.174 & 0.227 \\
\addlinespace[2pt]
SPIDER thorax & bAcc$\uparrow$ & \textbf{0.858} & 0.856 & 0.199 & 0.538 & 0.119 & 0.926 \\
 & ECE$\downarrow$ & 0.046 & \textbf{0.036} & 0.572 & 0.295 & 0.813 & 0.013 \\
 & NLL$\downarrow$ & 0.430 & \textbf{0.382} & 5.750 & 2.820 & 7.246 & 0.179 \\
\addlinespace[2pt]
SPIDER breast & bAcc$\uparrow$ & 0.659 & \textbf{0.718} & 0.195 & 0.380 & 0.145 & 0.860 \\
 & ECE$\downarrow$ & 0.094 & \textbf{0.063} & 0.587 & 0.403 & 0.508 & 0.014 \\
 & NLL$\downarrow$ & 0.913 & \textbf{0.765} & 4.893 & 3.913 & 4.770 & 0.264 \\
\addlinespace[2pt]
SPIDER skin & bAcc$\uparrow$ & 0.776 & \textbf{0.819} & 0.156 & 0.449 & 0.077 & 0.931 \\
 & ECE$\downarrow$ & \textbf{0.035} & 0.040 & 0.624 & 0.402 & 0.794 & 0.010 \\
 & NLL$\downarrow$ & 0.707 & \textbf{0.600} & 6.174 & 4.511 & 7.076 & 0.225 \\
\bottomrule
\end{tabular}
}{\textcolor{red}{[table pending]}}
\end{table}

\section{Implementation details}
\label{app:details}
\label{app:training}
This appendix gives the implementation details of the compared methods, the training budget of the two trained models, the pre-processing of the images, the CONCH prompts and what is known about the provenance of the data.

\paragraph{Compared methods.} \emph{OmniMed-Jev} is run with the 448-pixel pre-resize of its inference code (the processor then rescales to 896 pixels); because it often selects ``other'', it is also reported with that candidate removed and with one fixed random permutation of the candidates. \emph{Generative SFT}: every Choice, Noul and Score decision of a training state becomes one example, which gives \SFTExamplesPerPass{} examples. The step budget is pinned to PathJev-P's largest Stage~B budget (the ``5 epochs'' of the grid count passes over states); the selected checkpoint has made \SFTPassesAtCkpt{} passes over its examples, i.e.\ it has seen each decision about once, whereas the selected PathJev-P checkpoint has seen every state, with all its decisions, twice, with fewer optimiser steps (Table~\ref{tab:training}). At test time a candidate $c$ with tokens $t_1..t_n$ is scored by $s_c=\sum_{i}\log p(t_i\mid \text{image},\text{prompt},t_{<i})$, without an end-of-answer token and without length normalisation, and $p(c)=\mathrm{softmax}_c(s_c)$; no candidate text is a prefix of another in any test question. \emph{CONCH}: the L2-normalised text embeddings of the five prompts are averaged, and probabilities are the softmax of the model's learned logit scale times the cosine similarity with the projected image embedding. \emph{Linear probe}: the feature scaler, regularisation strength and temperature are chosen by grouped cross-validation on the training data; on HepatoBench and HiESD they are fitted, in each of five grouped outer folds, by inner grouped cross-validation on the outer training part only, with HiESD folds grouped by patient. HepatoBench images cannot be grouped by slide, so patches of one slide can fall on both sides of a fold there; regrouping in blocks of consecutive images changes the probe's balanced accuracy only slightly (Appendix~\ref{app:a3}). \emph{Inner-Jev}: probabilities are renormalised over the option letters; every letter is a single token, and the largest candidate set has 25 options.

\paragraph{Training budget.} Table~\ref{tab:training} gives, for the two trained methods, the selected configuration and the budget of the evaluated checkpoints. Both use NF4 4-bit weights with double quantisation and bf16 compute, LoRA on the same 476 decoder and 324 vision-encoder linear layers (the multimodal projector is not adapted), a paged 8-bit AdamW optimiser with linear schedule, 3\% warm-up, no weight decay and gradient clipping at 0.3, one GPU and 32 items per optimiser step. Marker tokens are existing single tokens of the vocabulary; their embeddings are not trained. No input was truncated. PathJev-P averages the cross-entropy over the decisions of a micro-batch; Gen-SFT averages the token-level cross-entropy over answer tokens (prompt and image masked), so its decisions are weighted by answer length.
\begin{table}[h]
\centering
\caption{Training budget of the evaluated checkpoints (from the run folders).}
\label{tab:training}
\scriptsize
\IfFileExists{generated/tab_training.tex}{
\begin{tabular}{@{}lll@{}}
\toprule
Quantity & PathJev-P (3 seeds) & Gen-SFT (3 seeds) \\
\midrule
Selected configuration & lr $2\times10^{-4}$, 3 epochs & lr $2\times10^{-4}$, budget of 5 PathJev-P epochs \\
Training items per pass & 40{,}000 states & 170{,}390 single-decision examples \\
Optimiser steps at checkpoint & 200 (A) + 2{,}500 (B) & 5{,}000 \\
Planned steps (whole run) & 3{,}950 & 6{,}250 \\
Items per optimiser step & 32 & 32 \\
Passes at checkpoint (Stage B) & 2.00 & 0.94 \\
Images seen up to checkpoint & 86{,}400 & 160{,}000 \\
Decision targets seen up to checkpoint & $\approx$368{,}042 & 160{,}000 \\
Trainable parameters & 39{,}555{,}202 & 38{,}497{,}792 \\
Wall time of the whole run (h, 1 GPU) & 10.4--10.5 & 16.3--16.5 \\
\bottomrule
\end{tabular}
}{\textcolor{red}{[table pending]}}
\end{table}

\paragraph{Pre-processing.} All MedGemma-based methods receive the whole image resized (square, aspect ratio not preserved) to $896\times896$ pixels, i.e.\ 256 visual tokens; for OmniMed-Jev the image is first resized to 448 pixels, as in its inference code. CONCH and the probe receive the whole image resized to $224\times224$. Native sizes differ by source: BRACS regions of interest are 237--5{,}324 pixels wide at 0.25\,$\mu$m/pixel (aspect ratio 0.42--2.58), SICAPv2 patches 512 pixels at $10\times$, HepatoBench 150 pixels at $20\times$, HiESD and SPIDER tiles 224 pixels at $20\times$. The field of view therefore varies across sources and, for BRACS, across images.

\paragraph{CONCH prompts.} ``an H\&E image of \{\}.'', ``a histopathology image showing \{\}.'', ``\{\}, H\&E
stain.'', ``a photomicrograph of \{\}.'', ``an image of \{\} tissue.'', with the candidate text in place of \{\}.

\paragraph{Provenance.} Table~\ref{tab:provenance} summarises what is documented about overlap between our datasets and the pretraining data of the models we use. No overlap is documented; for most cells it cannot be excluded either.
\begin{table}[h]
\centering
\caption{Overlap between datasets and model training data. D: documented; N: none documented and the model's fine-tuning data are listed; U: undisclosed training data, overlap cannot be excluded; R: dataset released after the model; the release date alone does not exclude overlap, because the images may predate it.}
\label{tab:provenance}
\scriptsize
\begin{tabular}{@{}lcccc@{}}
\toprule
Dataset & MedSigLIP / MedGemma 1.5 & OmniMed-Jev fine-tuning & CONCH v1 & H-optimus-0 \\
\midrule
SPIDER (HISTAI) & U & N & U & U \\
BRACS & U & N & U & U \\
SICAPv2 & U & N & U & U \\
HepatoBench & U & N & R & U \\
HiESD & U & N & R & U \\
\bottomrule
\end{tabular}
\end{table}

\section{Post-hoc robustness analyses (amendment A3)}
\label{app:a3}
These analyses were specified in amendment A3 of the analysis plan after the full results and an external review had been seen, and were frozen before they were run.
\paragraph{HepatoBench grouping (A3.1).} HepatoBench images are regrouped into blocks of 50 or 200 consecutive indices within each class, and H1a/H1b are re-tested exactly as in the primary analysis (Table~\ref{tab:a3hepato}); the comparisons are also repeated without HepatoBench. Because blocks are formed within each class, patches of one slide that belong to different classes stay in different blocks, and a block boundary can split a slide; the blocks are a sensitivity analysis, not a substitute for slide or patient groups. The probe, re-cross-fitted with block groups, reaches a HepatoBench balanced accuracy of \AThreeHepProbeBlockFiftyBacc{} (blocks of 50) and \AThreeHepProbeBlockTwoHundredBacc{} (blocks of 200), against \AThreeHepProbeImageBacc{} with image groups.
\begin{table}[h]
\centering
\caption{A3.1: H1a/H1b (clean macro, PathJev-P minus comparator, seed mean, 95\% marginal intervals) with different HepatoBench groups, and without HepatoBench.}
\label{tab:a3hepato}
\scriptsize
\IfFileExists{generated/tab_a3_hepato.tex}{
\begin{tabular}{@{}lcccccc@{}}
\toprule
& \multicolumn{2}{c}{H1a: $\Delta$ECE} & \multicolumn{2}{c}{H1b: $\Delta$bAcc} & \multicolumn{2}{c}{Supported} \\
\cmidrule(lr){2-3} \cmidrule(lr){4-5} \cmidrule(l){6-7}
HepatoBench groups & vs Gen-SFT & vs CONCH & vs Gen-SFT & vs CONCH & H1a & H1b \\
\midrule
Image (primary) & $-$0.075 & $-$0.199 & +0.049 & +0.196 & yes & yes \\
 & {\tiny[$-$0.088, $-$0.064]} & {\tiny[$-$0.237, $-$0.158]} & {\tiny[+0.038, +0.063]} & {\tiny[+0.161, +0.218]} & & \\
Blocks of 50 & $-$0.075 & $-$0.199 & +0.049 & +0.196 & yes & yes \\
 & {\tiny[$-$0.089, $-$0.063]} & {\tiny[$-$0.237, $-$0.156]} & {\tiny[+0.038, +0.063]} & {\tiny[+0.161, +0.218]} & & \\
Blocks of 200 & $-$0.075 & $-$0.199 & +0.049 & +0.196 & yes & yes \\
 & {\tiny[$-$0.091, $-$0.061]} & {\tiny[$-$0.238, $-$0.155]} & {\tiny[+0.038, +0.064]} & {\tiny[+0.161, +0.218]} & & \\
Without HepatoBench & $-$0.060 & $-$0.201 & +0.033 & +0.209 & yes & yes \\
 & {\tiny[$-$0.077, $-$0.045]} & {\tiny[$-$0.251, $-$0.144]} & {\tiny[+0.018, +0.052]} & {\tiny[+0.163, +0.237]} & & \\
\bottomrule
\end{tabular}
}{\textcolor{red}{[table pending]}}
\end{table}

\paragraph{Generative read-out (A3.4), candidate order (A3.5), OmniMed-Jev input size (A3.3) and blank images (A3.2).} Adding the log-probability of the end-of-answer token after each candidate changes candidate probabilities by at most $\AuditEosMaxDp{}$: the model gives that token a probability of at least \AuditEosMinProb{} after every candidate (exactly 1 in float32 for \AuditEosShareExact{}\% of candidates). Re-running the pre-registered read-out with the same checkpoint, inputs and code on L40 instead of H100 GPUs changed individual probabilities by up to \AuditRerunMaxDp{} and left the predicted class unchanged for at least \AuditRerunMinAgree{}\% of the states of each set; macro metrics moved by at most \AuditRerunMaxMacroDiff{}. The A3 tables use this re-run as the reference for the variants and break exact ties between candidates in a different order, which explains their small differences from Table~\ref{tab:main}.
\begin{table}[h]
\centering
\caption{A3.4: generative baseline with the pre-registered (summed) read-out of the candidate log-likelihoods, the same plus the end-of-answer token, and a length-normalised read-out, seed mean; main Choice question. First row: PathJev-P, from the runs of Table~\ref{tab:main}, for reference. Bold: best value per column.}
\label{tab:a3sftvar}
\scriptsize
\IfFileExists{generated/tab_a3_sftvar.tex}{
\begin{tabular}{@{}lccccccccc@{}}
\toprule
 & \multicolumn{3}{c}{Clean average} & \multicolumn{3}{c}{Unseen organs} & \multicolumn{3}{c}{In domain} \\
\cmidrule(lr){2-4} \cmidrule(lr){5-7} \cmidrule(l){8-10}
Read-out & bAcc$\uparrow$ & ECE$\downarrow$ & NLL$\downarrow$ & bAcc$\uparrow$ & ECE$\downarrow$ & NLL$\downarrow$ & bAcc$\uparrow$ & ECE$\downarrow$ & NLL$\downarrow$ \\
\midrule
PathJev-P (reference) & \textbf{0.541} & 0.204 & 1.586 & \textbf{0.390} & 0.301 & 2.394 & 0.692 & \textbf{0.108} & \textbf{0.777} \\
\midrule
Gen-SFT, sum (primary) & 0.492 & 0.279 & 2.257 & 0.276 & 0.446 & 3.700 & \textbf{0.709} & 0.112 & 0.814 \\
Gen-SFT, sum + end-of-answer & 0.492 & 0.279 & 2.257 & 0.276 & 0.446 & 3.700 & \textbf{0.709} & 0.112 & 0.814 \\
Gen-SFT, length-normalised & 0.492 & \textbf{0.201} & \textbf{1.508} & 0.289 & \textbf{0.110} & \textbf{1.965} & 0.696 & 0.293 & 1.050 \\
\bottomrule
\end{tabular}
}{\textcolor{red}{[table pending]}}
\end{table}
\begin{table}[h]
\centering
\caption{A3.5: random candidate orders on the clean sets. All methods use the same ten orders; Gen-SFT (seed 1 only) uses the first five. ``orig.'' is the stored run in the original order (Gen-SFT: seed 1, so it differs from the three-seed mean of Table~\ref{tab:main}); permuted runs used L40 GPUs and the stored runs H100, so part of the flips are numerical. TV: mean total-variation distance between a state's distribution under a permutation and under the original order; flips: share of states whose predicted candidate changes; ``other'': share of predictions of the ``other'' candidate, in the original order and over the permutations. Bold: best value per column.}
\label{tab:a3perm}
\scriptsize
\IfFileExists{generated/tab_a3_perm.tex}{
\setlength{\tabcolsep}{3.2pt}
\begin{tabular}{@{}lccccccccccc@{}}
\toprule
 & & \multicolumn{3}{c}{Clean bAcc$\uparrow$} & \multicolumn{3}{c}{Clean ECE$\downarrow$} & & & \multicolumn{2}{c}{``other'' (\%)} \\
\cmidrule(lr){3-5} \cmidrule(lr){6-8} \cmidrule(l){11-12}
Method & Perm. & orig. & mean & range & orig. & mean & range & TV$\downarrow$ & flips (\%)$\downarrow$ & orig. & perm. \\
\midrule
PathJev-P & 10 & 0.541 & \textbf{0.545} & 0.542--0.549 & 0.204 & \textbf{0.199} & 0.197--0.202 & \textbf{0.050} & \textbf{7.3} & 0.0 & 0.0 \\
Gen-SFT & 5 & 0.498 & 0.501 & 0.493--0.509 & 0.281 & 0.275 & 0.266--0.288 & 0.060 & 7.6 & 0.0 & 0.0 \\
OmniMed-Jev & 10 & 0.213 & 0.239 & 0.215--0.251 & 0.517 & 0.481 & 0.448--0.525 & 0.147 & 18.9 & 49.8 & 46.1 \\
\bottomrule
\end{tabular}
}{\textcolor{red}{[table pending]}}
\end{table}
\begin{table}[h]
\centering
\caption{A3.3: released OmniMed-Jev with images pre-resized to 448 (primary) or 896 pixels.}
\label{tab:a3omj}
\scriptsize
\IfFileExists{generated/tab_a3_omj896.tex}{
\begin{tabular}{@{}lcccccccc@{}}
\toprule
 & \multicolumn{4}{c}{Pre-resize 448 (primary)} & \multicolumn{4}{c}{Pre-resize 896} \\
\cmidrule(lr){2-5} \cmidrule(l){6-9}
Group & bAcc$\uparrow$ & ECE$\downarrow$ & NLL$\downarrow$ & Brier$\downarrow$ & bAcc$\uparrow$ & ECE$\downarrow$ & NLL$\downarrow$ & Brier$\downarrow$ \\
\midrule
Clean average & 0.213 & 0.517 & 3.269 & 1.185 & 0.207 & 0.519 & 3.291 & 1.191 \\
Unseen organs & 0.253 & 0.540 & 3.765 & 1.200 & 0.251 & 0.542 & 3.776 & 1.202 \\
In domain & 0.173 & 0.493 & 2.774 & 1.169 & 0.164 & 0.495 & 2.806 & 1.181 \\
Seen source & 0.212 & 0.577 & 5.261 & 1.269 & 0.210 & 0.581 & 5.348 & 1.277 \\
\bottomrule
\end{tabular}
}{\textcolor{red}{[table pending]}}
\end{table}
\begin{table}[h]
\centering
\caption{A3.2: 100 states per clean set with every image replaced by a uniform grey image. Max TV: largest total-variation distance between two outputs within the set. Constant predictions with balanced accuracy at chance show that, in this check, the constant text of the input does not separate the classes; they say nothing about pretraining contamination or patient independence.}
\label{tab:a3blank}
\scriptsize
\IfFileExists{generated/tab_a3_blank.tex}{
\begin{tabular}{@{}lcccccccc@{}}
\toprule
 & \multicolumn{2}{c}{HepatoBench} & \multicolumn{2}{c}{HiESD} & \multicolumn{2}{c}{BRACS} & \multicolumn{2}{c}{SICAPv2} \\
\cmidrule(lr){2-3} \cmidrule(lr){4-5} \cmidrule(lr){6-7} \cmidrule(lr){8-9}
Method & max TV & bAcc & max TV & bAcc & max TV & bAcc & max TV & bAcc \\
\midrule
PathJev-P & $3.6\times10^{-3}$ & 0.143 & $6.3\times10^{-3}$ & 0.100 & $3.6\times10^{-3}$ & 0.143 & $9.5\times10^{-4}$ & 0.250 \\
Gen-SFT & 0 & 0.143 & 0 & 0.100 & 0 & 0.143 & 0 & 0.250 \\
OmniMed-Jev & $1.1\times10^{-3}$ & 0.000 & $7.3\times10^{-4}$ & 0.100 & $1.3\times10^{-3}$ & 0.000 & $3.7\times10^{-4}$ & 0.000 \\
\midrule
\textit{Chance} & -- & 0.143 & -- & 0.100 & -- & 0.143 & -- & 0.250 \\
\bottomrule
\end{tabular}
}{\textcolor{red}{[table pending]}}
\end{table}

\section{Secondary and exploratory metrics}
\label{app:secondary}
This appendix reports the secondary metrics of the analysis plan, the calibration of every method before and after temperature scaling, and exploratory results on the Noul and Score questions.

\begin{table}[h]
\centering
\caption{Post hoc: Brier score on the unseen organs after one temperature per method, as in Table~\ref{tab:temperature}.}
\label{tab:temperaturebrier}
\scriptsize
\IfFileExists{generated/tab_temperature_brier.tex}{\begin{tabular}{@{}lccccc@{}}
\toprule
& \multicolumn{5}{c}{Temperature fitted on} \\
\cmidrule(l){2-6}
Method & none & val.\ mix & in-domain val. & other unseen organ & same set (oracle) \\
\midrule
PathJev-P (ours) & \textbf{0.843} & 0.790 & \textbf{0.772} & \textbf{0.723} & \textbf{0.715} \\
(b) Gen-SFT & 1.074 & 0.979 & 0.948 & 0.834 & 0.796 \\
(a) OmniMed-Jev & 1.200 & 0.821 & 0.863 & 0.818 & 0.817 \\
(c) CONCH & 1.094 & \textbf{0.778} & 0.880 & 0.773 & 0.777 \\
(e) Inner-Jev & 1.349 & 0.876 & 0.886 & 0.892 & 0.860 \\
\midrule
(d) Linear probe$^\dagger$ & 0.300 & 0.300 & 0.300 & 0.301 & 0.300 \\
\bottomrule
\end{tabular}
}{\textcolor{red}{[table pending]}}
\end{table}
\begin{table}[h]
\centering
\caption{Calibration per clean test set before (raw) and after a temperature fitted on the validation mix (scaled); seed mean for trained methods.}
\label{tab:scaled}
\scriptsize
\IfFileExists{generated/tab_scaled.tex}{\begin{tabular}{@{}llcccccc@{}}
\toprule
 & & \multicolumn{3}{c}{Raw} & \multicolumn{3}{c}{Validation temperature} \\
\cmidrule(lr){3-5} \cmidrule(l){6-8}
Set & Method & ECE$\downarrow$ & Brier$\downarrow$ & NLL$\downarrow$ & ECE$\downarrow$ & Brier$\downarrow$ & NLL$\downarrow$ \\
\midrule
HepatoBench & PathJev-P & \textbf{0.305} & \textbf{0.750} & \textbf{1.862} & 0.247 & \textbf{0.698} & 1.592 \\
 & Gen-SFT & 0.426 & 0.960 & 3.240 & 0.360 & 0.888 & 2.436 \\
 & OmniMed-Jev & 0.575 & 1.225 & 3.798 & 0.178 & 0.798 & 1.878 \\
 & CONCH & 0.496 & 1.091 & 3.911 & \textbf{0.077} & 0.760 & \textbf{1.577} \\
 & Inner-Jev & 0.525 & 1.160 & 3.671 & 0.144 & 0.859 & 2.023 \\
\cmidrule(l){2-8}
 & Probe$^\dagger$ & 0.007 & 0.045 & 0.098 & 0.010 & 0.046 & 0.099 \\
\midrule
HiESD & PathJev-P & \textbf{0.297} & \textbf{0.936} & \textbf{2.926} & 0.228 & 0.883 & 2.567 \\
 & Gen-SFT & 0.465 & 1.187 & 4.162 & 0.355 & 1.071 & 3.304 \\
 & OmniMed-Jev & 0.506 & 1.176 & 3.731 & 0.091 & 0.843 & 2.082 \\
 & CONCH & 0.478 & 1.096 & 4.904 & 0.068 & \textbf{0.795} & \textbf{1.960} \\
 & Inner-Jev & 0.746 & 1.539 & 6.450 & \textbf{0.013} & 0.893 & 2.269 \\
\cmidrule(l){2-8}
 & Probe$^\dagger$ & 0.034 & 0.555 & 1.180 & 0.025 & 0.553 & 1.180 \\
\midrule
BRACS & PathJev-P & \textbf{0.141} & \textbf{0.554} & \textbf{1.092} & 0.090 & 0.533 & 1.026 \\
 & Gen-SFT & 0.151 & 0.555 & 1.127 & 0.084 & \textbf{0.527} & \textbf{1.017} \\
 & OmniMed-Jev & 0.428 & 1.081 & 2.743 & \textbf{0.056} & 0.844 & 1.970 \\
 & CONCH & 0.393 & 0.885 & 3.089 & 0.058 & 0.681 & 1.464 \\
 & Inner-Jev & 0.585 & 1.266 & 3.981 & 0.099 & 0.859 & 2.019 \\
\cmidrule(l){2-8}
 & Probe$^\dagger$ & 0.113 & 0.530 & 1.057 & 0.096 & 0.523 & 1.038 \\
\midrule
SICAPv2 & PathJev-P & \textbf{0.075} & \textbf{0.230} & \textbf{0.462} & 0.052 & \textbf{0.221} & \textbf{0.411} \\
 & Gen-SFT & \textbf{0.075} & 0.241 & 0.499 & \textbf{0.046} & 0.231 & 0.424 \\
 & OmniMed-Jev & 0.557 & 1.257 & 2.806 & 0.224 & 0.859 & 1.732 \\
 & CONCH & 0.244 & 0.866 & 1.953 & 0.142 & 0.743 & 1.386 \\
 & Inner-Jev & 0.457 & 1.038 & 2.749 & 0.143 & 0.777 & 1.554 \\
\cmidrule(l){2-8}
 & Probe$^\dagger$ & 0.034 & 0.291 & 0.506 & 0.029 & 0.291 & 0.506 \\
\bottomrule
\end{tabular}
}{\textcolor{red}{[table pending]}}
\end{table}

Table~\ref{tab:secondary} reports the secondary metrics of the analysis plan for the main Choice question. Table~\ref{tab:noulscore} reports the Noul and Score decisions; these were not pre-registered as endpoints and are exploratory. The organ question is omitted because it is answerable from the question text.
\begin{table}[h]
\centering
\caption{Secondary metrics of the main Choice question, before temperature scaling; mean\,$\pm$\,SD over seeds for trained methods. $^\dagger$Cross-fitted within HepatoBench and HiESD (Section~\ref{sec:setup}).}
\label{tab:secondary}
\scriptsize
\IfFileExists{generated/tab_secondary.tex}{
\begin{tabular}{@{}llcccccc@{}}
\toprule
Set & Method & Acc$\uparrow$ & Macro-F1$\uparrow$ & aECE$\downarrow$ & Brier$\downarrow$ & NLL$\downarrow$ & AURC$\downarrow$ \\
\midrule
HepatoBench & PathJev-P & \textbf{0.512$_{\pm.022}$} & \textbf{0.459$_{\pm.043}$} & \textbf{0.305$_{\pm.017}$} & \textbf{0.750$_{\pm.030}$} & \textbf{1.862$_{\pm.186}$} & \textbf{0.254$_{\pm.011}$} \\
 & Gen-SFT & 0.414$_{\pm.041}$ & 0.338$_{\pm.058}$ & 0.426$_{\pm.026}$ & 0.960$_{\pm.063}$ & 3.240$_{\pm.282}$ & 0.375$_{\pm.054}$ \\
 & OmniMed-Jev & 0.267 & 0.255 & 0.575 & 1.225 & 3.798 & 0.518 \\
 & CONCH & 0.354 & 0.387 & 0.496 & 1.091 & 3.911 & 0.654 \\
 & Inner-Jev & 0.303 & 0.184 & 0.525 & 1.160 & 3.671 & 0.580 \\
\cmidrule(l){2-8}
 & Probe$^\dagger$ & 0.971 & 0.971 & 0.005 & 0.045 & 0.098 & 0.003 \\
\midrule
HiESD & PathJev-P & 0.267$_{\pm.009}$ & 0.208$_{\pm.002}$ & \textbf{0.297$_{\pm.026}$} & \textbf{0.936$_{\pm.023}$} & \textbf{2.926$_{\pm.049}$} & \textbf{0.547$_{\pm.012}$} \\
 & Gen-SFT & 0.140$_{\pm.011}$ & 0.101$_{\pm.016}$ & 0.465$_{\pm.029}$ & 1.187$_{\pm.026}$ & 4.162$_{\pm.081}$ & 0.802$_{\pm.015}$ \\
 & OmniMed-Jev & 0.239 & 0.185 & 0.506 & 1.176 & 3.731 & 0.685 \\
 & CONCH & \textbf{0.301} & \textbf{0.286} & 0.478 & 1.096 & 4.906 & 0.548 \\
 & Inner-Jev & 0.158 & 0.093 & 0.746 & 1.539 & 6.450 & 0.806 \\
\cmidrule(l){2-8}
 & Probe$^\dagger$ & 0.572 & 0.571 & 0.038 & 0.555 & 1.180 & 0.232 \\
\midrule
BRACS & PathJev-P & 0.582$_{\pm.040}$ & 0.565$_{\pm.062}$ & \textbf{0.140$_{\pm.037}$} & \textbf{0.554$_{\pm.046}$} & \textbf{1.092$_{\pm.042}$} & \textbf{0.223$_{\pm.038}$} \\
 & Gen-SFT & \textbf{0.598$_{\pm.024}$} & \textbf{0.600$_{\pm.029}$} & 0.150$_{\pm.031}$ & 0.555$_{\pm.014}$ & 1.127$_{\pm.033}$ & \textbf{0.223$_{\pm.010}$} \\
 & OmniMed-Jev & 0.216 & 0.191 & 0.426 & 1.081 & 2.743 & 0.693 \\
 & CONCH & 0.444 & 0.358 & 0.388 & 0.885 & 3.089 & 0.367 \\
 & Inner-Jev & 0.261 & 0.186 & 0.585 & 1.266 & 3.981 & 0.679 \\
\cmidrule(l){2-8}
 & Probe$^\dagger$ & 0.598 & 0.587 & 0.114 & 0.530 & 1.057 & 0.200 \\
\midrule
SICAPv2 & PathJev-P & \textbf{0.853$_{\pm.006}$} & \textbf{0.820$_{\pm.012}$} & 0.073$_{\pm.014}$ & \textbf{0.230$_{\pm.014}$} & \textbf{0.462$_{\pm.043}$} & \textbf{0.044$_{\pm.002}$} \\
 & Gen-SFT & 0.844$_{\pm.007}$ & 0.818$_{\pm.010}$ & \textbf{0.072$_{\pm.011}$} & 0.241$_{\pm.004}$ & 0.499$_{\pm.035}$ & 0.049$_{\pm.002}$ \\
 & OmniMed-Jev & 0.091 & 0.120 & 0.557 & 1.257 & 2.806 & 0.940 \\
 & CONCH & 0.261 & 0.192 & 0.244 & 0.866 & 1.954 & 0.743 \\
 & Inner-Jev & 0.385 & 0.270 & 0.457 & 1.038 & 2.749 & 0.632 \\
\cmidrule(l){2-8}
 & Probe$^\dagger$ & 0.793 & 0.761 & 0.035 & 0.291 & 0.506 & 0.074 \\
\midrule
Clean average & PathJev-P & \textbf{0.554$_{\pm.013}$} & \textbf{0.513$_{\pm.017}$} & \textbf{0.204$_{\pm.003}$} & \textbf{0.618$_{\pm.007}$} & \textbf{1.586$_{\pm.048}$} & \textbf{0.267$_{\pm.014}$} \\
 & Gen-SFT & 0.499$_{\pm.010}$ & 0.464$_{\pm.012}$ & 0.278$_{\pm.006}$ & 0.736$_{\pm.014}$ & 2.257$_{\pm.066}$ & 0.362$_{\pm.015}$ \\
 & OmniMed-Jev & 0.203 & 0.188 & 0.516 & 1.185 & 3.269 & 0.709 \\
 & CONCH & 0.340 & 0.306 & 0.402 & 0.984 & 3.465 & 0.578 \\
 & Inner-Jev & 0.277 & 0.183 & 0.578 & 1.251 & 4.213 & 0.674 \\
\cmidrule(l){2-8}
 & Probe$^\dagger$ & 0.734 & 0.722 & 0.048 & 0.355 & 0.710 & 0.127 \\
\midrule
Seen source & PathJev-P & 0.791$_{\pm.028}$ & 0.743$_{\pm.036}$ & 0.049$_{\pm.024}$ & 0.294$_{\pm.039}$ & 0.618$_{\pm.090}$ & 0.063$_{\pm.013}$ \\
 & Gen-SFT & \textbf{0.829$_{\pm.005}$} & \textbf{0.786$_{\pm.005}$} & \textbf{0.044$_{\pm.011}$} & \textbf{0.245$_{\pm.006}$} & \textbf{0.535$_{\pm.030}$} & \textbf{0.043$_{\pm.001}$} \\
 & OmniMed-Jev & 0.230 & 0.145 & 0.577 & 1.269 & 5.260 & 0.617 \\
 & CONCH & 0.458 & 0.369 & 0.393 & 0.896 & 4.104 & 0.354 \\
 & Inner-Jev & 0.146 & 0.075 & 0.688 & 1.486 & 6.316 & 0.802 \\
\cmidrule(l){2-8}
 & Probe$^\dagger$ & 0.926 & 0.903 & 0.010 & 0.111 & 0.224 & 0.011 \\
\bottomrule
\end{tabular}
}{\textcolor{red}{[table pending]}}
\end{table}
\begin{table}[h]
\centering
\caption{Exploratory (not pre-registered): Noul and Score decisions on the clean test sets, before temperature scaling; seed mean for trained methods. Only HiESD has a Score question among the unseen sets.}
\label{tab:noulscore}
\scriptsize
\IfFileExists{generated/tab_noul_score.tex}{
\begin{tabular}{@{}llcccc@{}}
\toprule
Question & Method & AUROC$\uparrow$ & ECE$\downarrow$ & NLL$\downarrow$ & Brier$\downarrow$ \\
\midrule
HepatoBench: tumour & PathJev-P & \textbf{0.893$_{\pm.005}$} & \textbf{0.103$_{\pm.031}$} & \textbf{0.498$_{\pm.115}$} & 0.132$_{\pm.024}$ \\
 & Gen-SFT & 0.883$_{\pm.023}$ & 0.163$_{\pm.038}$ & 0.757$_{\pm.205}$ & 0.180$_{\pm.033}$ \\
 & OmniMed-Jev & 0.734 & 0.107 & 0.593 & \textbf{0.117} \\
 & Inner-Jev & 0.707 & 0.787 & 4.438 & 0.781 \\
\midrule
HiESD: adenocarcinoma & PathJev-P & \textbf{0.897$_{\pm.005}$} & \textbf{0.086$_{\pm.007}$} & \textbf{0.463$_{\pm.028}$} & \textbf{0.127$_{\pm.004}$} \\
 & Gen-SFT & 0.882$_{\pm.014}$ & 0.142$_{\pm.010}$ & 0.726$_{\pm.047}$ & 0.166$_{\pm.010}$ \\
 & OmniMed-Jev & 0.865 & 0.170 & 0.842 & 0.190 \\
 & Inner-Jev & 0.759 & 0.440 & 1.911 & 0.456 \\
\midrule
HiESD: intestinal metaplasia & PathJev-P & \textbf{0.684$_{\pm.019}$} & \textbf{0.078$_{\pm.028}$} & \textbf{0.544$_{\pm.063}$} & \textbf{0.162$_{\pm.007}$} \\
 & Gen-SFT & 0.599$_{\pm.038}$ & 0.167$_{\pm.007}$ & 0.946$_{\pm.090}$ & 0.195$_{\pm.010}$ \\
 & OmniMed-Jev & 0.593 & 0.169 & 0.829 & 0.188 \\
 & Inner-Jev & 0.680 & 0.778 & 4.677 & 0.775 \\
\midrule
BRACS: carcinoma & PathJev-P & \textbf{0.940$_{\pm.002}$} & \textbf{0.069$_{\pm.006}$} & \textbf{0.357$_{\pm.052}$} & \textbf{0.093$_{\pm.002}$} \\
 & Gen-SFT & \textbf{0.940$_{\pm.001}$} & 0.081$_{\pm.001}$ & 0.373$_{\pm.026}$ & 0.098$_{\pm.001}$ \\
 & OmniMed-Jev & 0.756 & 0.217 & 0.863 & 0.227 \\
 & Inner-Jev & 0.898 & 0.333 & 1.530 & 0.355 \\
\midrule
BRACS: invasive carcinoma & PathJev-P & \textbf{0.998$_{\pm.001}$} & \textbf{0.011$_{\pm.000}$} & \textbf{0.044$_{\pm.003}$} & \textbf{0.013$_{\pm.001}$} \\
 & Gen-SFT & 0.996$_{\pm.002}$ & 0.023$_{\pm.007}$ & 0.095$_{\pm.021}$ & 0.023$_{\pm.007}$ \\
 & OmniMed-Jev & 0.782 & 0.088 & 0.458 & 0.108 \\
 & Inner-Jev & 0.946 & 0.539 & 3.080 & 0.549 \\
\midrule
SICAPv2: carcinoma & PathJev-P & \textbf{0.996$_{\pm.000}$} & 0.020$_{\pm.005}$ & \textbf{0.104$_{\pm.019}$} & \textbf{0.022$_{\pm.004}$} \\
 & Gen-SFT & \textbf{0.996$_{\pm.000}$} & \textbf{0.018$_{\pm.003}$} & 0.105$_{\pm.022}$ & 0.023$_{\pm.003}$ \\
 & OmniMed-Jev & 0.627 & 0.616 & 3.061 & 0.606 \\
 & Inner-Jev & 0.880 & 0.107 & 0.491 & 0.145 \\
\midrule
Unseen organs (average) & PathJev-P & \textbf{0.825$_{\pm.006}$} & \textbf{0.089$_{\pm.019}$} & \textbf{0.502$_{\pm.065}$} & \textbf{0.140$_{\pm.012}$} \\
 & Gen-SFT & 0.788$_{\pm.010}$ & 0.157$_{\pm.017}$ & 0.810$_{\pm.110}$ & 0.181$_{\pm.015}$ \\
 & OmniMed-Jev & 0.731 & 0.148 & 0.754 & 0.165 \\
 & Inner-Jev & 0.716 & 0.668 & 3.675 & 0.671 \\
\midrule
In domain (average) & PathJev-P & \textbf{0.978$_{\pm.001}$} & \textbf{0.034$_{\pm.003}$} & \textbf{0.169$_{\pm.024}$} & \textbf{0.043$_{\pm.001}$} \\
 & Gen-SFT & 0.977$_{\pm.001}$ & 0.041$_{\pm.004}$ & 0.191$_{\pm.016}$ & 0.048$_{\pm.003}$ \\
 & OmniMed-Jev & 0.722 & 0.307 & 1.461 & 0.314 \\
 & Inner-Jev & 0.908 & 0.326 & 1.700 & 0.350 \\
\bottomrule
\end{tabular}

\medskip

\begin{tabular}{@{}llccc@{}}
\toprule
Question & Method & MAE (levels)$\downarrow$ & RPS$\downarrow$ & NLL$\downarrow$ \\
\midrule
HiESD: correa stage & PathJev-P & \textbf{0.713$_{\pm.028}$} & \textbf{0.172$_{\pm.009}$} & \textbf{1.238$_{\pm.121}$} \\
 & Gen-SFT & 0.905$_{\pm.046}$ & 0.248$_{\pm.016}$ & 2.356$_{\pm.208}$ \\
 & OmniMed-Jev & 0.894 & 0.239 & 1.583 \\
 & Inner-Jev & 1.044 & 0.324 & 3.356 \\
\midrule
BRACS: risk group & PathJev-P & 0.318$_{\pm.016}$ & \textbf{0.103$_{\pm.001}$} & \textbf{0.638$_{\pm.054}$} \\
 & Gen-SFT & \textbf{0.310$_{\pm.015}$} & 0.114$_{\pm.009}$ & 0.747$_{\pm.111}$ \\
 & OmniMed-Jev & 0.561 & 0.216 & 1.363 \\
 & Inner-Jev & 0.615 & 0.297 & 2.700 \\
\midrule
SICAPv2: grade level & PathJev-P & \textbf{0.176$_{\pm.008}$} & \textbf{0.042$_{\pm.003}$} & \textbf{0.461$_{\pm.043}$} \\
 & Gen-SFT & 0.193$_{\pm.006}$ & 0.045$_{\pm.002}$ & 0.505$_{\pm.049}$ \\
 & OmniMed-Jev & 1.171 & 0.308 & 2.437 \\
 & Inner-Jev & 1.486 & 0.473 & 4.242 \\
\midrule
In domain (average) & PathJev-P & \textbf{0.247$_{\pm.005}$} & \textbf{0.073$_{\pm.002}$} & \textbf{0.550$_{\pm.045}$} \\
 & Gen-SFT & 0.252$_{\pm.005}$ & 0.080$_{\pm.005}$ & 0.626$_{\pm.078}$ \\
 & OmniMed-Jev & 0.866 & 0.262 & 1.900 \\
 & Inner-Jev & 1.050 & 0.385 & 3.471 \\
\bottomrule
\end{tabular}
}{\textcolor{red}{[table pending]}}
\end{table}

\section{Whole-slide demonstration: remaining slides}
\label{app:wsi}
This appendix shows the two slides of the whole-slide demonstration that are not shown in the main text.

\begin{figure}[h]
\centering
\includegraphics[width=0.8\textwidth]{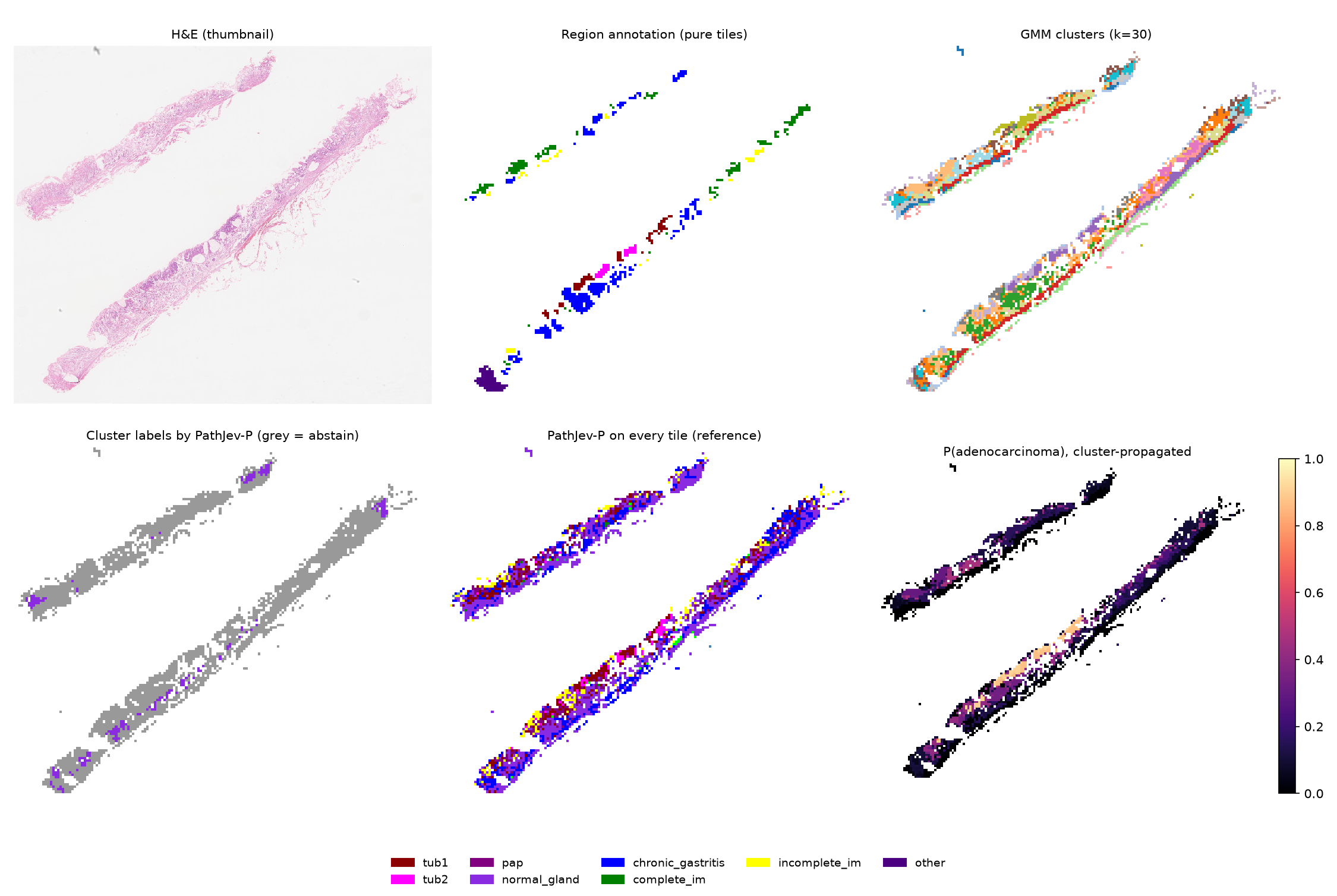}\\[2pt]
\includegraphics[width=0.8\textwidth]{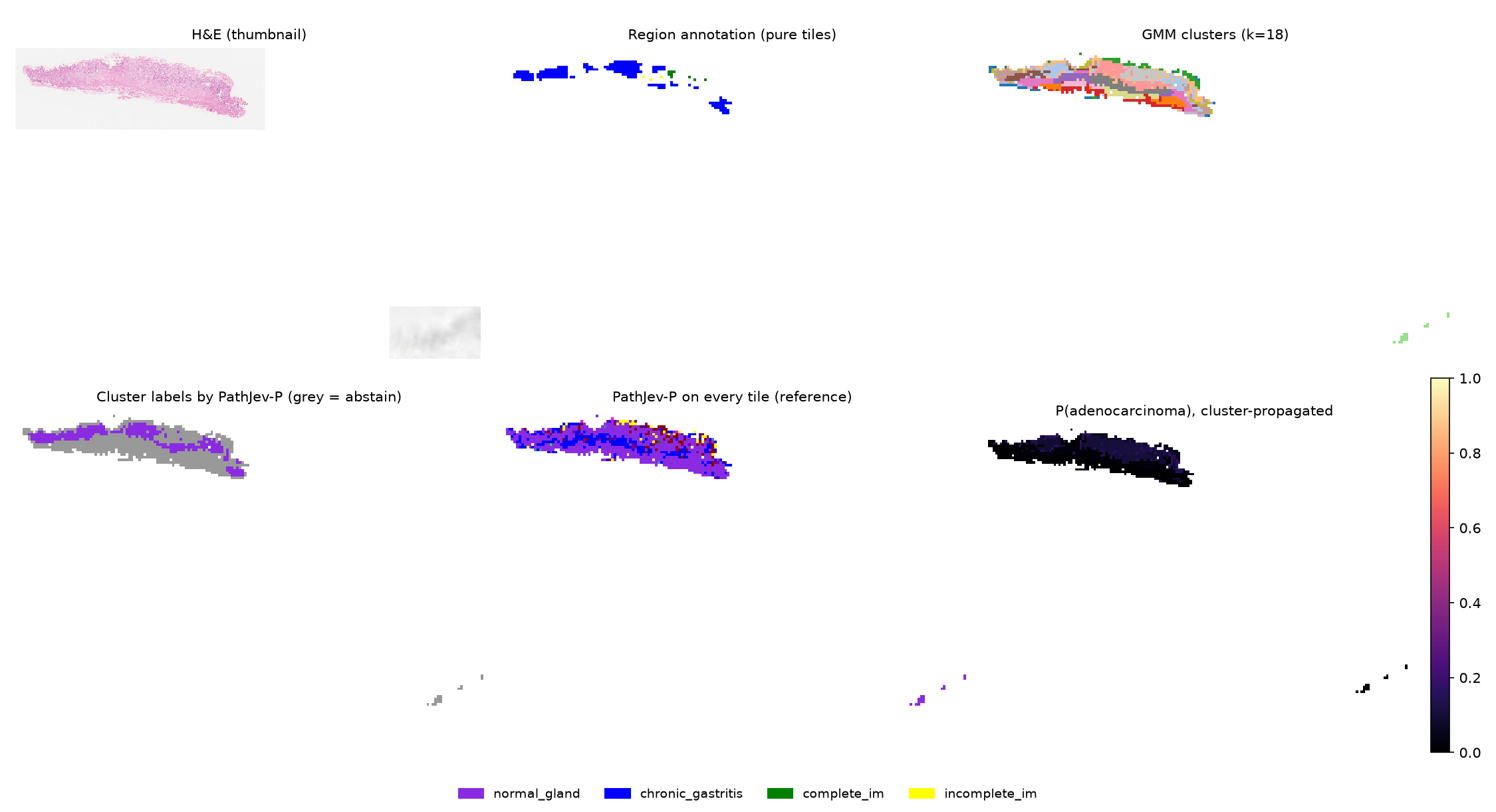}
\caption{The two HiESD slides of the demonstration not shown in Figure~\ref{fig:wsi} (top: slide with adenocarcinoma; bottom: slide without). Panels as in Figure~\ref{fig:wsi}.}
\label{fig:wsi-app}
\end{figure}
\end{document}